%% file: main.tex
\documentclass[10pt,twocolumn,letterpaper]{article}
\usepackage{scrextend}
\usepackage{cvpr}              % To produce the CAMERA-READY version
\usepackage{comment}
\usepackage{graphicx}
\usepackage{booktabs}
\usepackage{amsfonts}
\usepackage{algorithmic}
\usepackage{algorithm}
\usepackage{array}
\usepackage{textcomp}
\usepackage{stfloats}
\usepackage{url}
\usepackage{verbatim}
\usepackage{graphicx}
\usepackage{soul}
\usepackage{url}
\usepackage[utf8]{inputenc}
\usepackage{caption}
\usepackage{amssymb,amsmath,amsthm}

\usepackage{mathtools} % Bonus
\usepackage{booktabs}

\usepackage{amsthm}
\usepackage{tikz}
\usepackage{amsmath,amssymb} % define this before the line numbering.
\usepackage{color}

\usepackage[accsupp]{axessibility} 

\usepackage[pagebackref,breaklinks,colorlinks]{hyperref}

\usepackage[capitalize]{cleveref}
\crefname{section}{Sec.}{Secs.}
\Crefname{section}{Section}{Sections}
\Crefname{table}{Table}{Tables}
\crefname{table}{Tab.}{Tabs.}

\def\cvprPaperID{5867} % *** Enter the CVPR Paper ID here
\def\confName{CVPR}
\def\confYear{2023}

\begin{document}

%%%%%%%%% TITLE - PLEASE UPDATE
\title{Relighting Scenes with Object Insertions in Neural Radiance Fields}
\author{
    %Authors
    % All authors must be in the same font size and format.
   Xuening Zhu\thanks{Co-first authors.}, Renjiao Yi\footnotemark[1], Xin Wen, Chenyang Zhu, Kai Xu\thanks{Corresponding author.}\\
   %\footnotemark[1]   Co-first authors, \footnotemark[2]   Corresponding author. \\
   National University of Defense Technology\\
%{\tt\small yirenjiao, zhuchenyang07@nudt.edu.cn, kevin.kai.xu@gmail.com}
}

% \author[ author1, author2]{Qingmao},        %Qingmao属于第1，2个单位
% \author[author2]{Xiaosui},               %Xiaosui属于第2个单位
% \author[author3]{Houcheng},               %Houcheng属于第3个单位
% \author[author2]{Tonglin\corauthref{cor1}} %Tonglin属于第2个单位，同时还是通讯作者
% \corauthref[cor1]{Corresponding author.}
% \ead{zzz@163.com}
% \author{Renjiao Yi, \\
% Institution1\\
% Institution1 address\\
% {\tt\small firstauthor@i1.org}
% % For a paper whose authors are all at the same institution,
% % omit the following lines up until the closing ``}''.
% % Additional authors and addresses can be added with ``\and'',
% % just like the second author.
% % To save space, use either the email address or home page, not both
% \and
% Second Author\\
% Institution2\\
% First line of institution2 address\\
% {\tt\small secondauthor@i2.org}
% }

%%%%%%%%% TITLE - PLEASE UPDATE
% \title{Author Responses for Paper 5867}  % **** Enter the paper title here
% \title{Relighting Scenes with Object Insertions in Neural Radiance Fields}

% \author{\parbox{\textwidth}{\centering Xuening Zhu\thanks{Co-first authors.},  Renjiao Yi\footnotemark[1], Xin Wen, Chenyang Zhu, Kai Xu\thanks{Corresponding author.}
%         }
%         \\
% % For Computer Graphics Forum: Please use the abbreviation of your first name.
% {\parbox{\textwidth}{\centering $^1$National University of Defense Technology, China}}
% }

\twocolumn[{%
\maketitle
\begin{center}
\captionsetup{type=figure}
\includegraphics[width=1\linewidth]{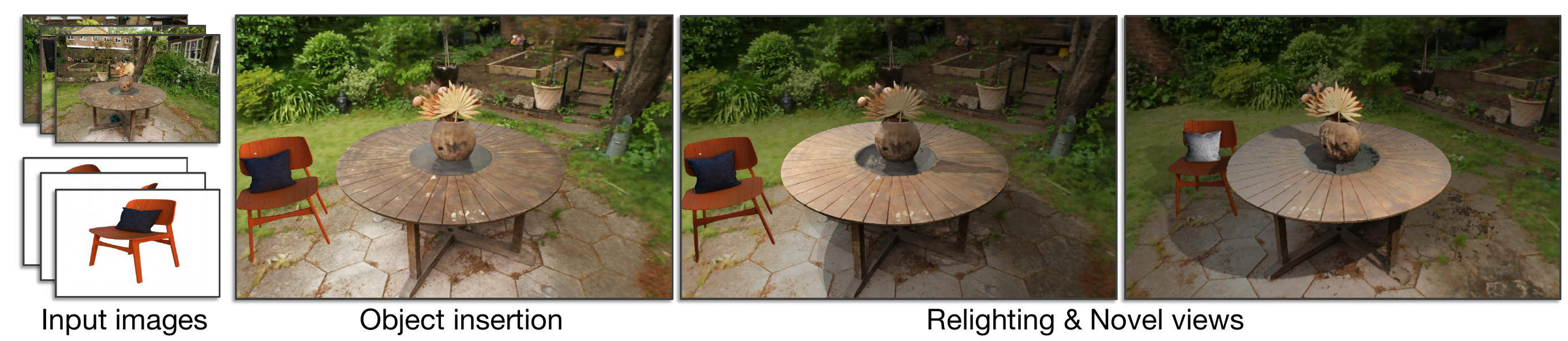}
% \vspace{-0.5em}
\captionof{figure}{\label{fig:teaser}We propose a method for inserting object NeRFs into scene NeRFs. From left to right, the figure depicts input images, object insertion without relighting, and relightings from novel views. The method also supports non-Lambertian relighting and material rendering, as the pillow on the chair.
}
\end{center}
}]
% \maketitle
\maketitle
\input{0_abstract}

\input{1_intro}
\input{2_related}

\input{3_method}
\input{4_experiment}

{\small
	\bibliographystyle{ieee_fullname}
	\bibliography{egbib}
}

\end{document}

%% file: 0_abstract.tex
\begin{abstract}
The insertion of objects into a scene and relighting are commonly utilized applications in augmented reality (AR). Previous methods focused on inserting virtual objects using CAD models or real objects from single-view images, resulting in highly limited AR application scenarios. We propose a novel NeRF-based pipeline for inserting object NeRFs into scene NeRFs, enabling novel view synthesis and realistic relighting, supporting physical interactions like casting shadows onto each other, from two sets of images depicting the object and scene. The lighting environment is in a hybrid representation of Spherical Harmonics and Spherical Gaussians, representing both high- and low-frequency lighting components very well, and supporting non-Lambertian surfaces. Specifically, we leverage the benefits of volume rendering and introduce an innovative approach for efficient shadow rendering by comparing the depth maps between the camera view and the light source view and generating vivid soft shadows. The proposed method achieves realistic relighting effects in extensive experimental evaluations.
%-------------------------------------------------------------------------
%  ACM CCS 1998
%  (see https://www.acm.org/publications/computing-classification-system/1998)
% \begin{classification} % according to https://www.acm.org/publications/computing-classification-system/1998
% \CCScat{Computer Graphics}{I.3.3}{Picture/Image Generation}{Line and curve generation}
% \end{classification}
%-------------------------------------------------------------------------
%  ACM CCS 2012
   % (see https://www.acm.org/publications/class-2012)
%The tool at \url{http://dl.acm.org/ccs.cfm} can be used to generate
% CCS codes.
%Example:
% \begin{CCSXML}
% <ccs2012>
%    <concept>
%        <concept_id>10010147.10010371.10010382.10010385</concept_id>
%        <concept_desc>Computing methodologies~Image-based rendering</concept_desc>
%        <concept_significance>500</concept_significance>
%        </concept>
%  </ccs2012>
% \end{CCSXML}

% \ccsdesc[500]{Computing methodologies~Image-based rendering}

% \printccsdesc   
\end{abstract}

%% file: 1_intro.tex
\section{Introduction}
\label{sec:intro}
%Object insertion has widespread applications across a variety of industries, including augmented reality (AR), virtual reality (VR), and the animation and film sectors. The aim of object insertion is to seamlessly incorporate objects into scenes and deliver immersive visual experiences. 

Object insertion is widely used across various industries, such as augmented reality (AR), virtual reality (VR), and the animation and film sectors. The goal of object insertion is to seamlessly integrate objects into scenes, enhancing visual immersion for users.
%that the user needs and to provide an immersive visual experience. 
%The composition and rendering of object and scene is a classic research 
%problem in computer graphics, which has been widely used in the film and 
%the television entertainment industry, VR, AR, and other fields. The goal of 
%the problem is to insert objects into the scene to generate the visual scene 
%that the user needs and to provide an immersive visual experience. 
Traditional computer graphics methods typically involve manually capturing and reconstructing the physical world, encompassing geometry, texture, and other attributes. The conventional industrial process often involves manually composing CAD models of scenes and objects, and using traditional rendering techniques like rasterization and ray tracing for relighting or generating new views, to create a customized visual experience. Nonetheless, this process is labor-intensive and not user-friendly for beginners.
%Traditional computer graphics approaches usually rely on the manual capture and reconstruction of the physical world, including geometry, texture, and other attributes. The standard industrial pipeline often entails manually compositing CAD models of scenes and objects, and employing traditional rendering methods such as rasterization and ray tracing for relighting or new view synthesis, to deliver a tailored visual experience. However, this pipeline is time-consuming and not user-friendly for amateurs. 

In this paper, we aim to introduce a thorough pipeline for compositing and relighting 3D objects and scenes solely based on image inputs. Addressing this challenge encompasses various highly ambiguous tasks, including 3D reconstruction, inverse rendering, and image-based rendering. NeRF~\cite{2020NeRF} provides a method that connects images and 3D scenes by utilizing MLPs and volume rendering to achieve realistic, high-quality effects in a straightforward manner. Therefore, we incorporate NeRF as the backbone of our proposed pipeline for end-to-end differentiable reconstruction and relighting. 
%In this paper, our goal is to present a comprehensive pipeline for compositing and relighting 3D objects and 3D scenes using only image inputs. This problem involves several highly ill-posed challenges, including 3D reconstruction, inverse rendering, and image-based rendering. NeRF~\cite{2020NeRF} offers a technique to bridge images and 3D scenes that leverages MLPs and volume rendering to deliver high-quality, realistic effects in a simple form. Therefore we adopt NeRF as the backbone of our proposed pipeline for end-to-end differentiable reconstruction and relighting.
%This method is novel in thought and simple in form and can achieve the amazing synthesis effect of a new perspective. 
%This method provides new ideas for rendering and view synthesis. However, 
%The original NeRF demonstrates significant potential as a novel implicit 3D representation. However, 

%A limitation of NeRF is the coupling of geometry, illumination, and material information in its representation. This inherent coupling poses challenges when attempting to directly apply NeRFs for composing and editing 3D objects and scenes. 
One drawback of NeRF is its inherent coupling of geometry, illumination, and material information within its representation. This interconnectedness presents challenges when trying to directly employ NeRFs for composing and editing 3D objects and scenes.
%The rendering method of new perspective synthesis proposed by NeRF can get high-quality pictures, but it is difficult to directly implement the objects inserted into the scene and generate a new perspective picture, the problem mainly includes the following aspects:
% In summary, 
Other challenges in applying NeRF for relighting include:
Firstly, NeRF simplifies volume rendering by assuming that particles are self-luminous, the light of particle radiation couples the light and material properties. Furthermore, the ambient light in the input image remains undetermined, impeding support for lighting editing. 
Secondly, there is a lack of a lighting rendering pipeline that accommodates both low- and high-frequency lighting sources, and supports spatially-variant lighting and non-Lambertian surfaces in NeRFs. 
%Other challenges of applying NeRF for relighting include:
%firstly, NeRF makes a simplified assumption for volume rendering that the particles are self-luminous, the light of particle radiation couples the light and material properties. Additionally, the ambient light in the input image remains unknown, hindering the support for lighting editing. 
%Secondly, there lacks a differentiable lighting rendering pipeline for both low- and high-frequency lighting sources and supports spatially-variant lighting, non-Lambertian surfaces in NeRFs. 
%Secondly, differences in sizes between scenes and objects pose a challenge, with scenes typically being larger and requiring a larger number of samples along the rays. Directly combining two NeRFs can result in missing samples or uneven sampling. 
Lastly, NeRF treats shadows as part of the color of objects, disregarding the visibility of incoming light sources, thus limiting its ability to produce realistic shadow effects.

To address the first challenge, we propose decoupling NeRF outputs into illumination-related and material-related terms, i.e. decomposing the color into reflectance and shading components. This approach effectively resolves the intrinsic decomposition of NeRFs in a completely unsupervised manner. With decomposed geometry, material, and lighting, we can further edit the scene by inserting object NeRFs into scene NeRFs and fitting them into the scene illumination by the proposed shading replacement, or relighting the whole composited scene. 

%Moving to the second challenge, we recognize the difference in sizes between scenes and objects. 
%In the coarse stage of NeRF sampling, 
%inadequate sampling points on inserted objects lead to inaccuracies in the probability density distribution function and potentially cause parts of the inserted object to be missing in rendered images. We introduce a segmented sampling strategy, which optimizes sampling distribution without a significant increase in computational costs.
%For the second challenge, many differentiable lighting rendering methods such as Pytorch3D use point lights, which can be computationally expensive for an environment map, since a large number of point lights would be needed for such low-frequency lighting. We introduce a hybrid representation of lighting by combining Spherical Harmonics and Spherical Gaussians, to represent low- and high-frequency components respectively. With rendering layers from \cite{yi2023weakly}, non-Lambertian surfaces can be rendered and materials of each point can be edited. 
Addressing the second challenge, many differentiable lighting rendering methods, such as Pytorch3D, utilize point lights. However, employing a large number of point lights for an environment map can be computationally intensive, particularly for low-frequency lighting. To overcome this, we introduce a hybrid lighting representation by combining Spherical Harmonics and Spherical Gaussians to respectively represent low- and high-frequency components. Leveraging rendering layers from \cite{yi2023weakly}, our approach enables the rendering of non-Lambertian surfaces and the editing of materials for each point. Spherical Gaussians are used to fill the gap between Spherical Harmonics lighting and original HDR environment maps. Furthermore, in our pipeline, except distant lighting as many prior works, we can also put Spherical Gaussian lighting at 3D positions, supporting spatially-variant lighting. 

Regarding the third challenge, previous studies have proposed two approaches for modeling light source visibility. One approach involves directly predicting the visibility of light sources for each point through neural networks but is limited to static scenes where scene geometry remains constant. The alternative approach generates shadow rays from each 3D point to the light source, using the density of sampling points on the shadow ray as the light source visibility coefficient for the 3D point. While this method is more suitable for composited or edited scenes, the extensive computation required to generate shadow rays significantly slows down shadow rendering. 
% In response, we introduce a novel and efficient strategy for shadow rendering that compares depth inconsistencies between the camera view and light view to swiftly compute light visibility at each point. 
In response, we introduce the variance shadow mapping strategy into volume rendering by comparing the depths from the light source's perspective, and Chebyshev's inequality to generate soft shadows.
This innovative approach delivers comparable results to conventional methods while substantially enhancing efficiency in both time and computation resources.

The whole pipeline is end-to-end and differentiable, in an unsupervised manner. The contributions are summarized as follows:
\begin{itemize} 
\item We propose a method to insert object NeRF into scene NeRF while fitting into the scene illumination, by solving the intrinsic decomposition of NeRFs unsupervisedly. 
%\item We propose a strategy of segmented sampling for compositing two NeRFs of different sizes, to solve the missing sampling of uniform sampling. 
\item We propose a pipeline to relight the composited scene, with a hybrid lighting representation of Spherical Harmonics and Spherical Gaussians, supporting non-Lambertian surfaces, cast shadows, and spatially variant light sources. 
\item We introduce an efficient way of variance shadow mapping in NeRFs, for realistic soft shadow rendering. %volume rendering of NeRFs, to generate realistic soft shadows efficiently. 
% generating shadows, combined with the idea of traditional computer graphics shadow mapping, and realize the effect of quickly adding shadows under the NeRF framework.
%modified the original uniform sampling method to segmented sampling, 
%The strategy does not increase many numbers of sampling points and avoids missing sampling as a result. %a large number of sampling points to ensure that the composition of different scenes with large different scales will not appear the loss of geometric details.
\end{itemize}
%-------------------------------------------------------------------------

%% file: 2_related.tex
\section{Related Works}

\begin{figure*}[tb]
      \centering
      \includegraphics[width=\linewidth]{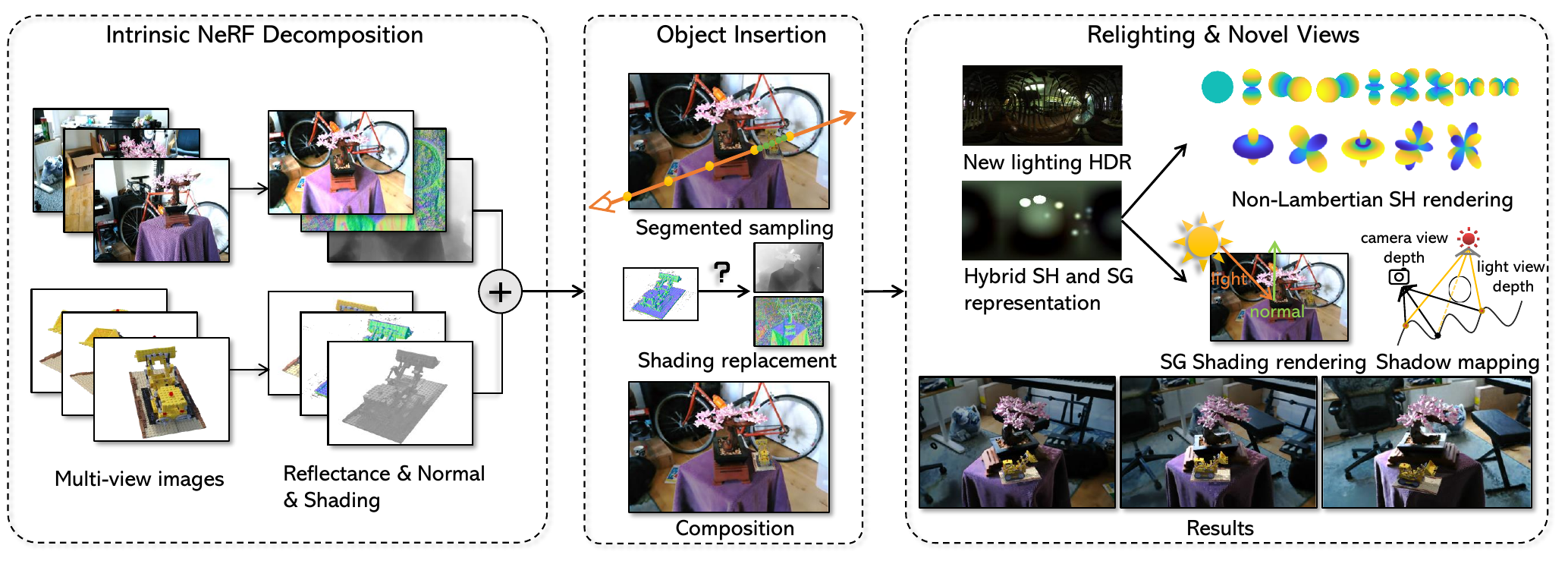}
      \caption{\label{fig:pipeline}%
            \textbf{Overview of the method.}
           Given two sets of multi-view images of a scene and an object from unknown lighting environments, our method reconstructs and predicts the intrinsic decomposition of the scene and object. Subsequently, it composites their geometry through segmented sampling, updating the object's shading to match the scene illumination. At last, non-Lambertian relighting of the composited scene is proposed with the hybrid lighting representation and efficient shadow mapping. }%The shading and shadow renderings are efficient by the proposed strategies. 
            %the segmented sampling and shading replacement strategies are introduced to achieve synthesis novel view of the composition of object and scene.
            %Furthermore, we utilize depth and normal to implement efficient rendering  with reasonable light and shadow effects of new lights.
                        
            %\vspace{-10px}
    \end{figure*}

\subsection{Inverse Rendering}

Before NeRFs, methods~\cite{yi2018faces, yu2023accidental} of object insertion usually insert virtual objects into the image, not supporting renderings from novel views. With inverse rendering, methods can re-render real objects from images into another scene~\cite{yi2023weakly}. The goal of inverse rendering is to extract the geometry, material properties, and lighting information of a scene from image inputs, for further manipulations such as relighting or material editing. Classical methods typically depend on optical devices to measure various physical factors, including geometric structures.
%, reflectance, and environmental illumination of the object or scene. 
Neural networks have recently emerged as a solution for tasks like scene relighting and inverse rendering. The representative CGIntrinsic \cite{CGIntrinsics} decomposes the reflectance and grayscale shading of the scene by learning and realizes approximate lighting editing by modifying Shading. However, it is limited to altering the lighting of static images and does not facilitate the generation of new perspective views of the modified scene.
%Recently, many light-based decoupling methods from NeRF have been introduced. 
NeRF \cite{2020NeRF} achieves high-quality new perspective synthesis effects by representing scenes with implicit MLP, and a large number of related improvements \cite{2020DeRF, 2021BARF, Niemeyer20arxiv_GIRAFFE, 2021Mip, 2022DEFORMABLE, Ye_2023_CVPR} have emerged since then.
Some NeRF-based methods \cite{NeuralBRDF, 2020Neural, 2020NeRD, 2021Neural, 2021NeRFactor, 2021NeRV, 2021PhySG, 2021osr, choi2023iblnerf, boss2022-samurai, 2022NeILF, NeILF++}, model the BRDF, illumination intensity, and illumination visibility separately to achieve inverse rendering. 
Since directly decomposing these attributes from a single static scene is an ill-posed problem, prior knowledge, such as known illumination or input images depicting different lighting scenarios, is commonly incorporated.

Zeng $et$ $al$.~\cite{zeng2023nrhints} introduce a relightable radiance field that utilizes known light sources as inputs to achieve accurate highlights and shadows. Lyu $et$ $al$.~\cite{lyu2022nrtf} follow the idea of PRT to represent the transmission results of lighting and use Mitsuba to add OLAT images as supervision for inverse rendering.
IntrinsicNeRF~\cite{ye2023intrinsicnerf} achieves intrinsic image decomposition in NeRFs. These papers focus on the inverse rendering process and relighting of the scene, without changing the scene by adding or deleting objects. %with more focus on changing the material of the object and the attempted effect of different illumination. 
Our paper focuses on the composition of scene and object, where inverse rendering is adopted to support the relighting of NeRF, for realistic composition and AR results. It supports spatially variant light sources and cast shadow rendering in relighting. % is adopted two different lighting scenes and objects and produces reasonable light and shadow effects.

\subsection{Scene Composition in NeRFs}
%The composition of objects and scenes based on the NeRF 
% According to the content of composition, these

Related methods~\cite{2022Control, yang2021objectnerf, kobayashi2022decomposing, tang2022compressiblecomposable, 2020Object, wang2023neural, wang2023udcnerf, wu2022object, Wu2023objectsdfplus} can be roughly divided into the composition of inputs of MLPs and the composition of color and voxel density. Control-NeRF~\cite{2022Control} is a typical method for compositing inputs, they propose to separate the scene representation and the network to learn a potential scene feature representation. 
%In this way, the network is no longer a representation of a single scene. To edit the scene, they simply edit the input feature vector. Multiple scenes can generalize from the same network to obtain the 3D scene representations. Therefore, to insert the object into the scene, they modify the feature vector input, as the sum of feature vectors of the corresponding points of the object and the scene. Then the composited color and voxel density are predicted through the network. 
%Control-NeRF is the most similar work to ours. 
This method improves the generalization of NeRF and can quickly achieve scene composition, but it does not consider the impact of illumination. 
%For scenes and objects with illumination gaps, it cannot generate a reasonable visual effect. 
%Another category is to composite the voxel density and color of the corresponding points. 
%They usually separate a single object from the scene and then move the object to other places in the scene. 
%Guo $et$ $al$. \cite{2020Object} realize the composition of several objects.
%, but it is not suitable for the composition of objects and scenes.
Yang $et$ $al$.~\cite{yang2021objectnerf} propose two MLP branches to model the background and object properties respectively to separate individual objects from the scene. 
%, and different objects are distinguished by different learnable encodings
Sosuke $et$ $al$.~\cite{kobayashi2022decomposing} adopt one branch for predicting the color of voxels and the other branch to predict semantic features, separating individual objects by semantic separation. 
Tang $et$ $al$.~\cite{tang2022compressiblecomposable} learn a hybrid tensor rank decomposition of the scene, which can be arbitrarily composited into one scene by concatenating along the rank dimension. These methods focus more on how to separate individual objects from complex scenes. They edit objects that are originally from the scene, where they do not have to consider the consistency of illumination and often ignore the rationality of post-composition shadows.
%Wang $et$ $al$. \cite{wang2023neural} realize the direct insertion of virtual objects, where they have known 3D models. 
Different from these methods, the goal of our paper is to realize object insertion into scenes, fitting into the scene illumination, and support further relighting. % with spatially variant light sources and cast shadows. % of the composition of objects and scenes from different environments, and further relighting the scene. 
%by adding new lights.
% \vspace{-10px}

\subsection{Shadow Mapping}
Shadow mapping is a conventional technique for rendering shadows in computer graphics, and the advancement of technology has led to the development of numerous shadow mapping algorithms. 
Initially, shadow mapping~\cite{shadowmapping} compares the depth of the light source view to determine the presence of shadows, resulting in distinct hard shadows. 
To address the issue of soft shadows, many methods have been developed~\cite{pcss, pcf, VSM, csm, msm}.
The Percentage-Closer soft shadow~\cite{pcf} method incorporates multiple depths and blends them to generate soft shadows. 
The variance shadow mapping method~\cite{VSM} utilizes depth variance to achieve soft shading effects through the application of Chebyshev inequality.
Additionally, some approaches utilize neural networks for fractional shadow calculations, as demonstrated by references such as Karnieli $et$ $al$. and others~\cite{karnieli2022deepshadow,10.1007/978-3-031-19827-4_18}. These methods employ neural networks to learn depth and generate shadows, often represented as scenes produced through supervised learning techniques. 
In this paper, we introduce Variance Shadow Mapping (VSM) into Neural Radiance Fields (NeRF) for end-to-end relighting.

%% file: 3_method.tex
%-------------------------------------------------------------------------

\section{Method}
As presented in Fig.~\ref{fig:pipeline}, given two sets of posed images of a scene and an object, $\{I_{si}, C_{si}\}_{i=1}^{N_s}$ and $\{I_{oi}, C_{oi}\}_{i=1}^{N_o}$, both captured in an unknown lighting environment. 
%$\{I_{si}, C_{si}\}_{i=1}^{N_s}$, and a set of posed images of an object captured in another unknown lighting environment $\{I_{oi}, C_{oi}\}_{i=1}^{N_o}$. 
$I$ is the RGB image and $C$ is the camera pose. 
We aim to insert the object into the scene. 
%After compositing the geometry of the scene and object, we need to ensure that the lighting is consistent. Furthermore, we aim to relight the composited scene by spatially variant lighting with vivid soft shadows. % to render images of the scene and objects.
% In this section, we first introduce some preliminaries about NeRF briefly in Section~\ref{sectionpre}. 
The method includes three parts. The first part is to how to obtain the intrinsic NeRFs, which is introduced in Section \ref{section1}. 
The second part is the object insertion into the scene, which is introduced in Section \ref{section2}. The third part is relighting of the composited scene, as introduced in Section~\ref{sec:relighting}. 
%-------------------------------------------------------------------------

% \subsection{Preliminary of NeRFs}
% \label{sectionpre}
% %\subsubsection*{Neural Radiance Fields.}
% NeRF \cite{2020NeRF}uses an MLP to model a continuous neural radiation field: $F(X, D)=(\sigma,c)$.
% The inputs of the MLP are 3D location $X \in R^3$ and direction $D \in R^3$ (both inputs are in positional encoding). The outputs are the volume density $\sigma$ and RGB color $c$.
% After obtaining this implicit representation of the 3D scene, the image from any new view can be rendered according to the volume rendering formula:
% \vspace{-6px}
% \begin{equation}
%       \label{volume-redering}
%       \begin{split}
%       \hat{\mathbf{C}}(\mathbf{r})=\int_{t_{n}}^{t_{f}} T(t) \sigma(t) \mathbf{c}(t) d t, \\
%       T(t)=\exp \left(-\int_{t_{n}}^{t_{f}} \sigma(s) d s\right), \\
%       \end{split}
% \end{equation}
% where $\mathbf{r}$ is the camera ray with the origin at $\mathbf{o}$ and the direction of $\mathbf{d}$, which can be represented as ${\mathbf{r}(t)=\mathbf{o}+t\mathbf{d}}$. $t_{n}$ and $t_{f}$ are the nearest and farthest distance along the ray $\mathbf{r}$. Our pipeline is based on this original NeRF backbone. 

%------------------------------------------------------

\subsection{Intrinsic NeRF Decomposition}
\label{section1}
\subsubsection{Intrinsic NeRF Model}
%In the real world, most lights received by human eyes are from diffuse reflection. 
%Therefore, only the diffuse reflection of scenes and objects is considered in this paper.
To address the highly ill-posed problem of inverse rendering 3D scenes without any supervision, we utilize a simplified model known as intrinsic image decomposition. 
% Specifically, intrinsic image decomposition is a long-studied vision problem where an image is decomposed into a reflectance image and a shading image, which are element-wise products. 
% We extend this problem to NeRF, upon which our intrinsic NeRF ${F_I}$ is based.
Our model has three outputs: density $\sigma$, reflectance $\mathbf{R} \in R^3$ and shading $\mathbf{S} \in R$, i.e. ${F_I(X, D)=(\sigma,\mathbf{R},\mathbf{S})}$.
While reflectance is solely dependent on the object's properties, we only use the 3D locations as input for the reflectance MLP. On the other hand, shading is influenced by the location and viewing perspective. Therefore, the inputs for the shading MLP include the location and view direction. 
Without any additional input, we only use the RGB images for supervision:
\begin{equation}\label{INeRF}
\begin{split}
&\hat{\mathbf{C}}(\mathbf{r})=\int_{t_{n}}^{t_{f}} T(t) \sigma(t)  \big(\mathbf{S}(t) \cdot \mathbf{R}(t)\big) d t ,\\
&\mathcal{L}=\frac{1}{|\mathcal{R}|}\sum_{\mathbf{r} \in \mathcal{R}}\|\hat{\mathbf{C}}(\mathbf{r})-\mathbf{C}(\mathbf{r})\|_{2}^{2},
\end{split}
\end{equation}
where $\mathbf{C}(\mathbf{r})$ is the ground truth RGB, $\mathcal{R}$ denotes the set of camera rays in a single batch.

\subsubsection{Decomposition of Geometry}
\begin{figure}
      \centering
       \includegraphics[width=\linewidth]{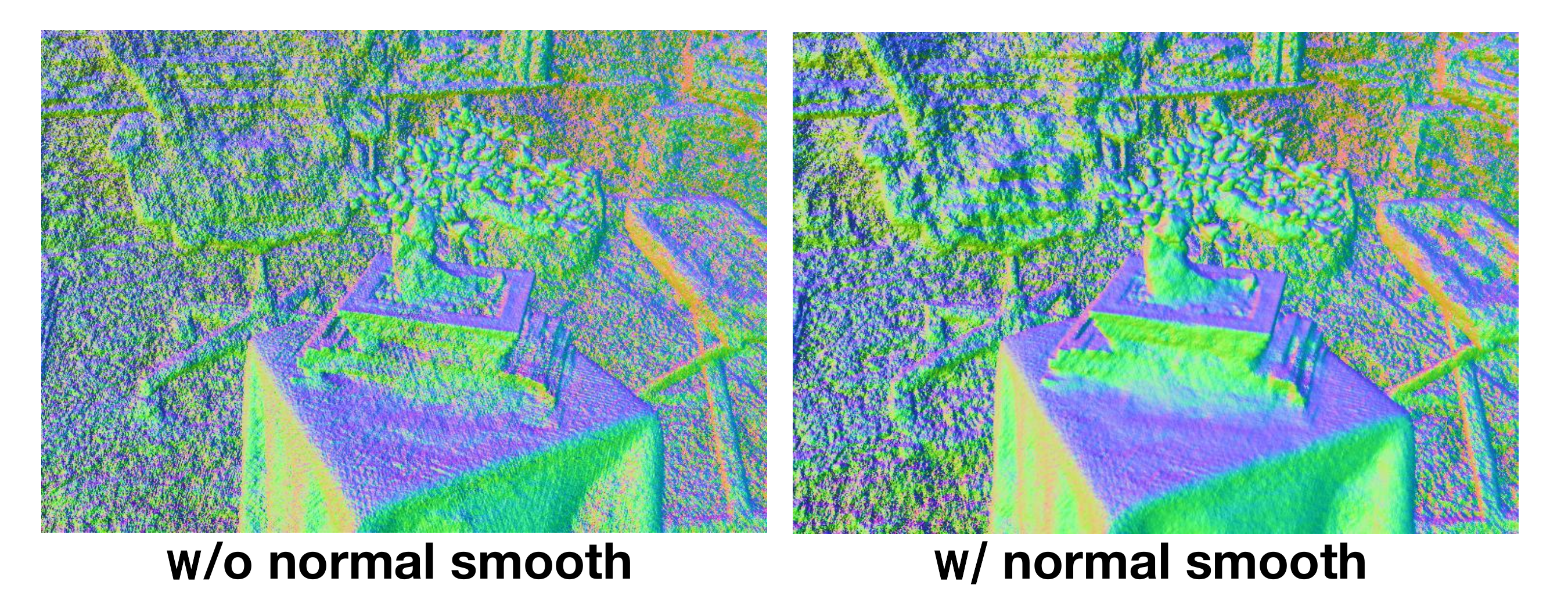}  
      \caption{\label{fig:normal}%
      Visualizations comparing cases with (on the right) and without (on the left) the term $\mathcal{L}_d$. This row displays illustrative examples from a scene. 
      }
      \vspace{-6mm}
\end{figure}

Given an arbitrary view and the intrinsic neural radiance field, normal and depth maps of the view can be computed easily:
% Specifically, the depth map $\mathbf{D}$ is obtained by integrating the point's depth along the ray, while normal map $\mathbf{N}$ is computed by integrating the point's normal direction along the ray. The normal direction of a point is derived from the negative normalized gradient of the $\sigma$-density of the intrinsic NeRF:
\begin{equation}\label{Geo}
\begin{split}
&\mathbf{D}(\mathbf{r})=\int_{t_{n}}^{t_{f}} T(t) \sigma(t) t d t, \\
&\mathbf{N}(\mathbf{r})=\int_{t_{n}}^{t_{f}} T(t) \sigma(t) \big(-\nabla(\sigma)\big) d t .
\end{split}
\end{equation}
% However, the raw normal map obtained in this manner exhibits significant noise, as depicted in the first column of Fig.~\ref{fig:normal}. 
% It is observed that the surface normal of adjacent points typically demonstrates consistency and continuity. 
To enhance the precision and smoothness of geometry for improved subsequent rendering tasks, we introduce a regularization term for normal directions to ensure consistency among adjacent points.
% Specifically, each batch takes $1024$ pixels, with the first $512$ pixels being randomly sampled and the left $512$ pixels being the neighboring pixels corresponding to the previous pixels. 
% The neighboring pixels of $\mathbf{r}$ are denoted as $\mathbf{r}_{adj}$, while the non-neighboring pixels of $\mathbf{r}$ are denoted as $\mathbf{r'}$. 
% Due to projective transformations of cameras, 
Following the pinhole camera model, a pixel in the background corresponds to a larger physical area. To guide the regularization process, we introduce a depth-dependent weight represented as:
\begin{equation}\label{LDepth}
\mathcal{L}_d=\frac{1}{|\mathcal{R}|}\sum_{\mathbf{r} \in \mathcal{R}}\big(1-\frac{\mathbf{N}(\mathbf{r})\cdot \mathbf{N}(\mathbf{r}_{adj})}{|\mathbf{N}(\mathbf{r})||\mathbf{N}(\mathbf{r}_{adj})|}\big)\frac{1}{\mathbf{D}(\mathbf{r})^2},
\end{equation}
where $\mathbf{N}(\mathbf{r})$ is the normal of pixel $\mathbf{r}$, $\mathbf{N}(\mathbf{r}_{adj})$ is the normal of the neighborhood around pixel $\mathbf{r}$. 
The comparison is shown in Fig.~\ref{fig:normal}. It is evident that the normal maps in the first column (without smoothness regularization) exhibit significantly more noise, whereas the normal maps in the second column with smoothness regularization appear smoother. This regularization is crucial, particularly for complex scenes.

\subsubsection{Decomposition of Reflectance and Shading}
% Decomposing reflectance and shading from images is an ill-posed problem, and prior assumptions are often introduced to solve it.
Reflectance is an inherent attribute of surfaces and represents the material properties of objects and scenes, independent of illumination. In traditional intrinsic image decomposition, constraints of reflectance chromaticities $\mathcal{L}_{chro}$, sparsity $\mathcal{L}_{rs}$ and non-local sparsity $\mathcal{L}_{nrs}$ are commonly used for optimization. 
% For example, in a recent work IntrinsicNeRF \cite{ye2023intrinsicnerf}, priors including chromaticity prior $\mathcal{L}_{chro}$, reflectance sparsity $\mathcal{L}_{rs}$ and non-local reflectance sparsity $\mathcal{L}_{rns}$ are considered. 
Here we introduce similar priors into the NeRF-based backbone. 

Additionally, a scale ambiguity exists between shading and reflectance in intrinsic image decomposition, where scaling factors of k on reflectance and 1/k on shading do not impact the observed image. To overcome unreasonable brightness changes in the reflectance, we constrain the consistency of reflectance and brightness of different points. It ensures that the shading of new objects matches the brightness level of the scene shading when inserted into the scene, avoiding excessive brightness or darkness.
%In addition, they consider
% we can also compare different views, where reflectance is consistent among views, and shading is changing. 
%from observations, we find the reason for pixel intensity variations among views is lighting changes, rather than reflectance changes. 
% Therefore, to enforce the reflectance to be invariant among views, %make the change of scene intensity more reflected in the change of shading, 
% we control the overall brightness of the reflectance.
Specifically, a mean reflectance regularization is used to constrain the reflectance of different scenes at the same level:
\begin{equation}\label{Lavg_r}
\begin{split}
\mathcal{L}_{mean\_r}=\frac{1}{|\mathcal{R}|}\sum_{\mathbf{r} \in \mathcal{R}}\Big|\big(\mathbf{R}(\mathbf{r}).max\big)-\beta_{avg\_r}\Big|,
\end{split}
\end{equation}
where $\beta_{avg\_r}$ is a hyperparameter, and it is set as 0.6 for most scenes. $\mathbf{R}(\mathbf{r}).max$ represents the highest value of reflectance in the RGB channels.

% \subsubsection{Decomposition of Shading}
Shading 
%obtained from our Intrinsic Neural Radiance Field 
represents the brightness of the light reflected from one point to the viewpoint and conveys information about environmental illumination. 
% To separate shading and reflectance during the unsupervised decomposition process, regularization constraints are introduced.
%It is common to find that most of the brightness in the scene changes little and is quite uniform. 
% Firstly, shading usually changes smoothly in the scene. we add a regularization to constrain the smoothness of shading throughout the scene. 
The shading smoothness $\mathcal{L}_s$ is defined as the variance of shading for all the pixels in a single batch.
Since shading and surface normal are correlated, there exists a consistency between shading and normal:
\begin{small}
\begin{equation}\label{L_sn}
      \begin{split}
      &\mathcal{L}_{sn}=\frac{1}{|\mathcal{R}|}\sum_{\mathbf{r} \in \mathcal{R}}\omega_n(\mathbf{r},\mathbf{r'})*|\mathbf{S}(\mathbf{r})-\mathbf{S}(\mathbf{r'})|*(1-tanh(\mathbf{Dis}(\mathbf{r}, \mathbf{r'})),\\
      % &\omega_n(\mathbf{r_1},\mathbf{r_2})=1+\frac{\mathbf{N}(\mathbf{r_1})\cdot \mathbf{N}(\mathbf{r_2})}{|\mathbf{N}(\mathbf{r_1})||\mathbf{N}(\mathbf{r_2})|}, \\
    \end{split}
\end{equation}
\end{small}
where $\omega_n$ measures the cosine similarity of $\mathbf{r}$ and $\mathbf{r'}$, $\mathbf{Dis}(\mathbf{r}, \mathbf{r'})$ represents the distance between their respective 3D points. 
%where $\omega_n$ measures the cosine similarity of $\mathbf{r}$ and $\mathbf{r'}$, $\mathbf{Dis}(\mathbf{r}, \mathbf{r'})$ represents the distance between their respective 3D points. 
% If the two normal vectors are more similar, the shading should also be more similar.

The intrinsic NeRF decomposition is trained by summing all the losses: %, as shown in the equation:
\begin{equation}\label{trainloss}
\begin{split}
\mathcal{L}_{train}=& \lambda_0\mathcal{L}+\lambda_1\mathcal{L}_d+\lambda_2\mathcal{L}_{mean\_r}+\lambda_3\mathcal{L}_{chro}\\
& +\lambda_4\mathcal{L}_{rs}+\lambda_5\mathcal{L}_{nrs}+\lambda_6\mathcal{L}_s+\lambda_7\mathcal{L}_{sn}, \\
\end{split}
\end{equation}

where all the $\lambda$ are adjustable weights.

%-------------------------------------------------------------------------
\subsection{The Composition of Scene and Object}
\label{section2}
Once the intrinsic NeRFs of the scene and the object are obtained respectively, we can insert objects into scenes based on geometry and shading. This composited scene can then be rendered from different views for new view synthesis, similar to the original NeRF approach.

%Next, we will introduce how to realize the new perspective synthesis of the composition of objects and scenes with the trained Intrinsic Neural Radiance Fields.

\subsubsection{Geometry Composition by Segmented Sampling}
\begin{figure}
      \centering
        \includegraphics[width=\linewidth]{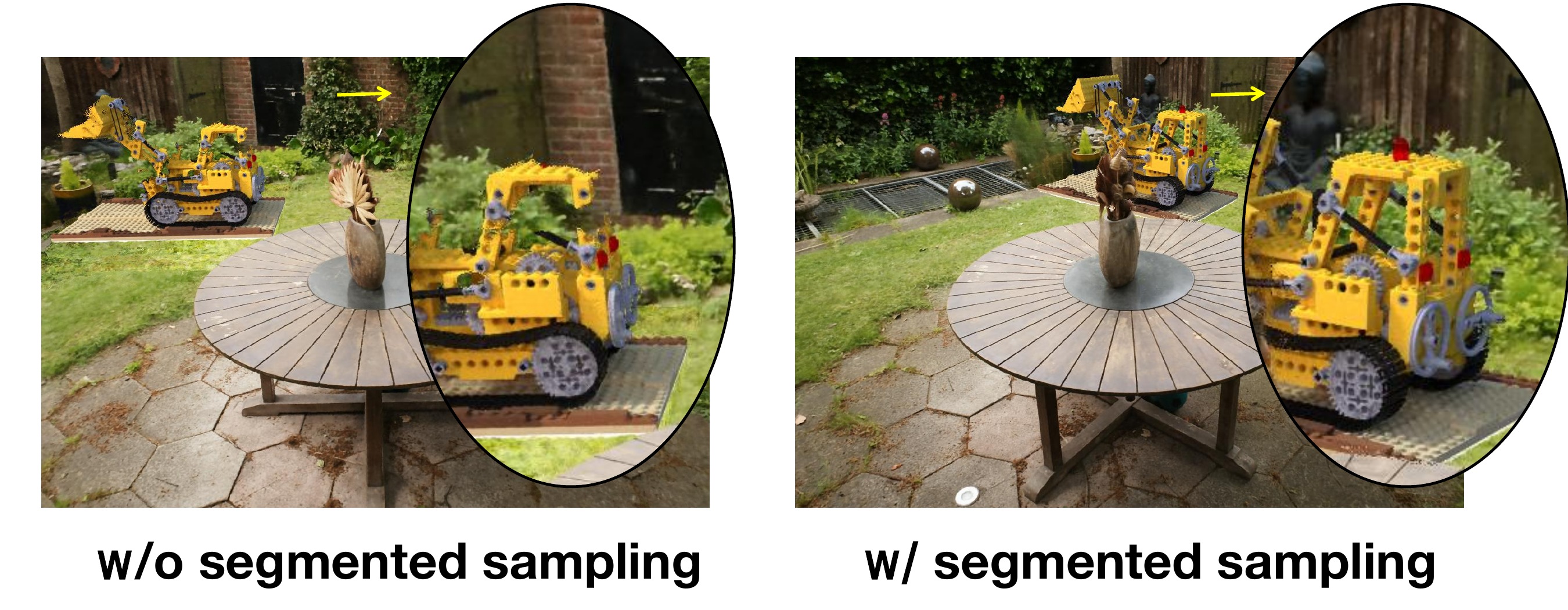} 
      \caption{\label{fig:geo}%
      Visualization of the rendered images depicting the segmented sampling strategy. Part of the Lego is missing on the left image. }
      %The left figure shows the result of using the original NeRF sampling strategy, which is uniform sampling. The enlarged area clearly shows that the inserted Lego lacks some geometric structure.
      %The figure on the right shows the results of using the segmented sampling strategy, with the geometric information of the Lego being fully preserved. The rendering time costs are comparable. }
      % \vspace{-12px}
      
\end{figure}

Given the inserting position and orientation in the scene (i.e. the translation and rotation between the coordinate systems of the object and scene), 
sampling points in the scene can be transformed to obtain the point coordinates in the object's coordinate system.
The $\sigma$-density, reflectance, and shading of the corresponding points are composited separately.
The composited density is the normalized sum of densities of corresponding points from the scene and the object at the same position. % of the corresponding points' density.
The reflectance is the weighted sum of reflectance values of corresponding points, where the weights are the densities of the voxel particles. Shading can be composited similarly if not considering relighting. 
%When composing shading without considering lighting consistency, the process is the same as compositing reflectance. % method for shading is the same as reflection, then 
The final image can be rendered following Eqn. \ref{INeRF}. 

However, during the experiment, it was observed that direct composition could lead to missing areas of the object in the rendered images, as shown in Fig.~\ref{fig:geo}. This issue arises due to insufficient sampling in the object area. Since scenes are typically larger than objects, the distance between adjacent sampling points in the scene is uniform. The sampling points on the object may be sparse because the object NeRF's size is smaller than the scene.
As shown in Fig.~\ref{fig:geo} (left), when sampling points are sparse on the object, it is possible that some parts of the object fall between two sampled points, and completely missing in the rendered images. It will lead to inaccurate probability density distribution (PDF) in the fine stage of sampling of NeRFs. This problem is very common in experiments. To solve the problem, a new sampling strategy called segmented sampling is proposed. This method selectively increases the sampling rate only within the bounding box of the object area to ensure detailed sampling without significantly increasing the overall number of sampling points and computational cost. This segmented sampling approach aims to capture the object details accurately while maintaining efficiency in the rendering process.
%\vspace{-20px}
\subsubsection{Relighting the Object into Scene Illumination}
The illumination of different scenes and objects usually varies greatly, and a naive composition without considering lighting consistency would lead to unrealistic results. %introduced above does not take into account the difference in the illumination. 
It is crucial to update the shading to ensure a realistic integration.  %the new light situation is unrelated to the original object shading. 
However, as the scene illumination is unknown, re-rendering shading from scratch is not feasible. We propose a shading replacement method to quickly generate object shading under the scene's illumination. %Therefore, we need to relight the object and replace the shading.
For Lambertian scenes, points with similar surface normals tend to have similar shading. Therefore, we can replace the shading of each object point with the shading of a nearby scene point that shares a similar normal vector. By finding these matching pairs and averaging their shading values, we can simulate the object shading under the scene illumination without direct knowledge of the lighting conditions. %where there are similar normals in the scene, simulating the relighting process of the object.
Specifically, for each object point, we can find scene points of similar normal vectors to form a pair. Since many scene points are similar in normal, there are many pairs for an object point. These pairs are ranked according to the normal similarities and distances between the object point and the scene points, and the mean shading value of the top-ranked ones $\mathbf{S}_{mean}$ is taken as the re-rendered shading value of the object point. 
In addition, the resulting shading needs to complement the difference in reflectance brightness by $\mathbf{S}_{updated}=\mathbf{S}_{mean}\frac{\beta_{avg\_r}^s}{\beta_{avg\_r}^o}$,
$\beta_{avg\_r}^s$ and $\beta_{avg\_r}^o$ are the average brightness of reflectance of the scene and the objects respectively. 
Fig.~\ref{fig:ablatio1} shows that by this way, inserting the object around the edge between a bright and a dark region gives a reasonable shading. 
% To achieve a more realistic rendering effect and further render the cast shadows, we infer one dominant light source from the shading to render shadows. 
% %It is represented as a point light to simulate the dominant light that cast shadows. 
% The updated shading of the object can be expressed as $\mathbf{S}_{updated}= I* \frac{\mathbf{N}(\mathbf{r})\cdot \mathbf{L}(l)}{|\mathbf{N}(\mathbf{r})||\mathbf{L}(l)|}$. The position of the light source can be obtained by solving the overdetermined equation with the formula. With the location of the light source, we can render cast shadows following Sec. \ref{sec:Shadow}.

\subsection{Relighting the Composited Scene}
\label{sec:relighting}

{In this section, we introduce how to render the composited scene under new HDR lighting.} 

\subsubsection{Hybrid Parametric Lighting Model}
Parametric models including Spherical Harmonics (SH) and Spherical Gaussians (SG) are commonly used for lighting representation. While second-order SH bases model the whole scene by only 9 coefficients for white illumination, and 27 coefficients for color illumination, and the simplified computation of rendering from SH coefficients as in \cite{ramamoorthi2001efficient}, it is commonly adopted in differentiable pipelines such as \cite{yu2019inverserendernet}. However, Spherical Harmonics cannot represent high-frequency lighting sources such as point lights or small area lights. Spherical Gaussians, on the contrary, represent high-frequency lighting sources very well but have difficulties representing low-frequency lighting with limited parameters. It is natural to combine them as a hybrid representation of lighting HDR. 

{To fit Spherical Gaussians on a given HDR, we first identify the three brightest points in the HDR image and record their pixel coordinates. Subsequently, these coordinates are transformed into spherical coordinates as the center/mean of three SGs. %, i.e. mean of SGs. % the unit sphere coordinate system to precisely determine the positions of the three primary light sources. 
%Then, we set the parameters of the spherical Gaussian: the mean corresponds to the positions of the primary light sources, and 
To enforce SGs to fit small area lights, we fix the variance. 
The variance constitutes a diagonal matrix with all diagonal elements set to 0.005. Small SGs can fit high-frequency lighting for precise shadow rendering. Finally, we project the Spherical Gaussians onto the spherical panorama, %, resulting in the derivation of the lighting map of Spherical Gaussians. 
and the difference between the lighting HDR and the lighting map of Spherical Gaussians is employed for fitting Spherical Harmonic lighting. We fit the second-order Spherical Harmonics for each channel, and there are 27 coefficients in total. SH lighting fits low-frequency lighting in the original HDR. }

Examples of the hybrid representation of Spherical Harmonics and Spherical Gaussians are shown in Figure~\ref{fig:sh+sg}, by representing the original HDR as low-order Spherical Harmonics and several Spherical Gaussians, the rendered shading and shadows are similar to ground truths. We set the number of Spherical Gaussians as 3 in the pipeline. As shown in Figure~\ref{fig:evaluteRelighting}, the renderings of the proposed pipeline are similar to the renderings of Blender. We can effectively render non-Lambertian surfaces, as well as precise cast shadows. The rendering step is detailed below. As lighting is addable, we can add the rendered images of SH and SGs to get the final rendering.

\subsubsection{Non-Lambertian Rendering of Spherical Harmonics}

Normal maps can be obtained from NeRFs by Eqn.~\ref{Geo} for a new view. 
We adopt the diffuse rendering layer according to the Phong model, each ray $(\mathbf{r})$ corresponds to a pixel on the normal map: 

\begin{equation}
\centering
\begin{aligned}
	%I(p)=&I_d(p)+I_s(p),\\
	I_d(\mathbf{r})=&\mathcal{R}(\mathbf{r})S(\mathbf{r})=\mathcal{R}(\mathbf{r})\sum_{\omega \in \mathcal{L}}l_\omega(L_\omega\cdot N(\mathbf{r})),
\label{equation:phong}
\end{aligned}
\end{equation}
%\textcolor{red}{add equation 2: $I_d=...$ from cvpr paper}, 
$\mathcal{R}(\mathbf{r})$ denotes the reflectance, and lighting can be replaced by SH representation as: 
%\textcolor{red}{add equation 12: $I_d=...$ from cvpr paper}. 
\begin{equation}
	\mathcal{L}=\sum_{l=0}^{\infty}\sum_{m=-l}^{l}C_{l,m}Y_{l,m}, \label{equation:SH}
\end{equation}
\noindent $Y_{l,m}$ is the SH basis of degree $l$ and order $m$, $C_{l,m}$ is the corresponding coefficient. Here we use up to the second order SH bases, i.e. $m \leq 2$, which are 9 bases in total. 

We adopt the rendering process in \cite{ramamoorthi2001relationship}, where Eqn.~\ref{equation:phong} becomes: 
\begin{equation}
	\label{equation:sh2}
%\begin{align}
	I_d(\mathbf{r})=\mathcal{R}(\mathbf{r})\sum_{\omega \in \mathcal{L}}l_\omega(L_\omega\cdot N(\mathbf{r}))=\mathcal{R}(\mathbf{r})\sum_{l,m}\hat{A}_l C_{l,m}Y_{l,m}(\theta,\phi),
		%S =\sum_{l,m}\hat{A}_l C_{l,m}Y_{l,m}= C \cdot F, \\
		%		I = R \odot (C\cdot F),
%\end{align}
\end{equation} 

We also adopt the specular rendering layer from \cite{yi2023weakly} to render the specular reflections by:

\begin{equation}
\begin{aligned}
	I_s(\mathbf{r})&=s_p\sum_{\omega \in \mathcal{L}}l_\omega(\frac{L_\omega+v}{\|L_\omega+v\|}\cdot N(\mathbf{r}))^\alpha \\
	&\approx s_p\sum_{l,m} C_{l,m}(\hat{A}_l\hat{Y}_{l,m}(\theta,\phi))^\alpha.
\end{aligned}\label{equation:sp}
\end{equation}

\noindent $\hat{Y}_{l,m}(\theta,\phi)$ are specular SH basis defined in \cite{yi2023weakly}. $s_p$ and $\alpha$ are material parameters controlling the specular reflectance and glossiness. By editing these parameters for each pixel, we can edit the material of each part of the composited scene. By adding $I_d$ and $I_s$, we get the rendered image of Spherical Harmonic lighting. 

\subsubsection{Shading Rendering of Spherical Gaussians}

%\textcolor{red}{to update} %In addition to the original scene illumination, we can also add or completely change the light sources of the scene. 
%Take the point light source as an example, 
%We can render new shading and compute the new images of the composited scene from any perspective illuminated by spatially variant lighting. 
%given the location and intensity of the new light source. %  if the location information and intensity of a light source are known.
We render each SG as 5 point lights, where the first point light is the SG center and others are sampled around the center with intensities diminishing according to Gaussian probabilities. Here if the lighting environment is a HDR, we can assume distant lighting, i.e. put SGs far from the scene. 
We can also model spatially variant lighting if the lights are put at 3D positions in the scene, with the light intensity diminishing as the distance of propagation increases. %.  can be effectively modeled using multiple point light sources at different 3D positions, with the light intensity diminishing as the distance of propagation increases.  
Shading $k$ point lights with intensity $\mathbf{I}_i$ is rendered by:
\begin{equation}\label{New_shading}
      \begin{split}
            \mathbf{S}_{SG}(\mathbf{r})=\sum_{i=1}^k\frac{\mathbf{N}(\mathbf{r})\cdot \mathbf{L}_i(\mathbf{r})}{|\mathbf{N}(\mathbf{r})||\mathbf{L}_i(\mathbf{r})|}*\frac{\mathbf{I}_i}{\gamma \cdot{\mathbf{Dis}_i(\mathbf{r})^2}},
      \end{split}
\end{equation}
where $\mathbf{L}_i(\mathbf{r})$ is the $i$th new light's direction, which is calculated from the position of the light source to the target point. If the lights are in the scene, i.e. they are spatially variant, then their directions vary from ray to ray. $\gamma$ is a hyperparameter and $\mathbf{Dis}_i(\mathbf{r})$ is the distance between light and the 3D surface points corresponding to the ray $\mathbf{r}$. They are used to control the luminance of the light. Lights would decay as the distance to the light source gets larger, respecting the real lighting effects in the real world. 
%The new shading is then added to or replaces the original shading, according to the relighting effects required. 
%We can further get the new images by multiplying shading to reflectance. If we want to completely change the lighting environments, we can simply replace the original shading with the rendered one. 
%A new lighting environment can be represented as a combination of point lights by Monte Carlo sampling on the lighting sphere. 

\begin{figure}[tb]
      \centering
\includegraphics[width=0.98\linewidth]{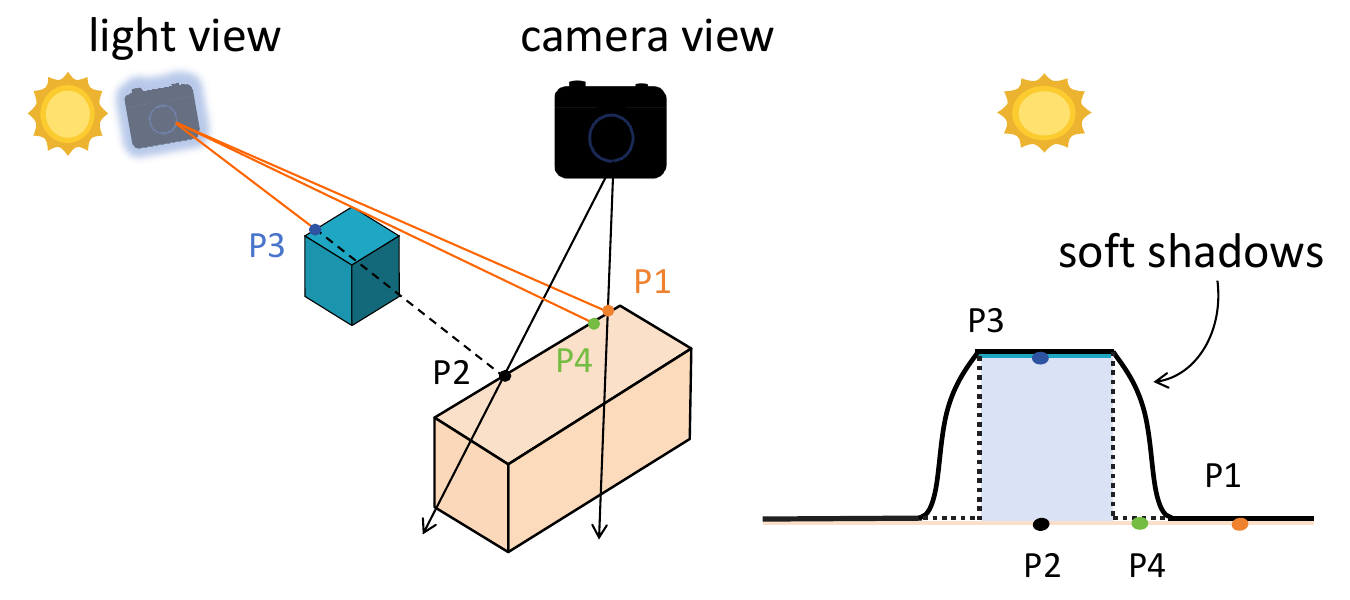}   
      \caption{\label{fig:shadow}%
        Variance shadow mapping. By comparing depth inconsistencies from the light view and the camera view, we have efficient shadow mapping for rendering cast shadows. For example, depths of P1 and P4 are consistent under two views, then they are visible. Depths of P2 are inconsistent in two views, so it is occluded, i.e. in shadows. In the right, blue areas are hard shadows by naive shadow mapping, where P4 is not in shadows. By variance shadow mapping, P4 is in soft shadows with visibility in $[0,1]$. }
        % \vspace{-10px}
\end{figure}
%\vspace{-5px}
%NEW SHADOW. 

\subsubsection{Shadow Rendering}
\label{sec:Shadow}
The shadow effect is a crucial aspect of rendering, enhancing the realism of images by simulating the occlusion of light by objects. %When introducing new light sources, it is essential to render the corresponding shadows to achieve a more authentic visual result.
We use SGs to render cast shadows. We are inspired by the variance shadow-mapping ideas \cite{VSM} from traditional computer graphics and propose a shadow rendering method fitting NeRFs.
% We are inspired by the shadow-mapping ideas from \cite{1978Casting} from traditional computer graphics, and propose a shadow rendering method in NeRF. 
% The shadow map of any viewing perspective can be represented as the visibility of the light source of that perspective.
Specifically, according to the depth map and camera parameters, the coordinates of surface points in the 3D space can be obtained. We set a virtual camera at the light source and transform the depth map from the camera view to this new light view to obtain $z(\mathbf{r})$. 
We also render the square map of the depth, denoted as $\mathbf{X^2}$, and the depth map $\mathbf{X}$ directly from the light view. 
These two maps are subjected to box filtering, resulting in the filtered versions denoted as $\mu(\mathbf{X^2})$ and $\mu(\mathbf{X})$. Lastly, the probability of pixel $\mathbf{r}$ not being occluded, i.e. the visibility, can be expressed as:

\begin{equation}\label{shadow_equ}
% \begin{centering}
    \begin{split}
       \operatorname{V(\mathbf{r})}&=\left\{
            \begin{array}{cc}1, & z(\mathbf{r}) \leqslant \mu(\mathbf{X})(\mathbf{r_l}) \\
            \frac{\sigma^2(\mathbf{r_l}) }{\sigma^2(\mathbf{r_l}) +(z(\mathbf{r})-\mu(\mathbf{X})(\mathbf{r_l}) )^2}, & z(\mathbf{r})>\mu(\mathbf{X})(\mathbf{r_l}) 
            \end{array}\right. , \\
        &\sigma^2(\mathbf{r_l})=\mu(\mathbf{X^2})(\mathbf{r_l})-\mu^2(\mathbf{X})(\mathbf{r_l}) \\
    \end{split}
    % \end{centering}
\end{equation}
where $\mathbf{r_l}$ is the correspondence pixel of $\mathbf{r}$ in light view.
In Fig.~\ref{fig:shadow}, the left side illustrates the key insight of shadow mapping, which involves comparing the depths of the points to be rendered from the viewpoint of the light source. The right side demonstrates how variance shadow mapping transforms the discontinuous depth viewed from the perspective of the light source (represented by the black dashed line) into a continuous depth (illustrated by the black solid line), leading to a final shadow in a continuous range in $[0,1]$.

% By comparing the depth consistency between depth maps computed from camera view and rendered directly, we can infer the visibility of points.  
% As illustrated in Fig.~\ref{fig:shadow}, based on the depth map rendered from the camera view and camera parameters, we can calculate the coordinates of surface points $P1$ and $P2$. Given the camera parameters of the virtual camera at the light source, we can convert $P1$ and $P2$ to the perspective of the light source to obtain the depth of the corresponding pixel of the light source view. Since $P2$ is occluded, the depth maps are not consistent. The depth value at this point in light view is smaller. In this case, the visibility of $P2$ is set as 0. Similarly for unoccluded points such as $P1$, the depth values are consistent, and the visibility is set as 1.
This method only requires generating depth maps from two views, avoiding the cumulative transmittance of the light along the sampling point on the shadow ray.
This approach works efficiently in practice. At last, we can get the final rendering for each ray $\mathbf{r}$ by:
\begin{equation}
I(\mathbf{r}) = I_d(\mathbf{r})+I_s(\mathbf{r})+V(\mathbf{r})\mathcal{R}(\mathbf{r})S_{SG}(\mathbf{r}). % for each ray $\mathbf(r)$. %$\textcolor{red}
\end{equation}
%{$S_{newlight}$}. 
%-------------------------------------------------------------------------

%% file: 4_experiment.tex
% \vspace{-6px}
\section{Experiments}
We conduct experiments for intrinsic decomposition, ablations, and relighting. %Please refer to supplementary videos for more results. %video results. \textit{Supplementary} for additional results, including more ablations, applications, and relighting videos.
%\vspace{-6px}

\begin{figure*}
\vspace{-5px}
      \centering
      \includegraphics[width=0.98\linewidth]{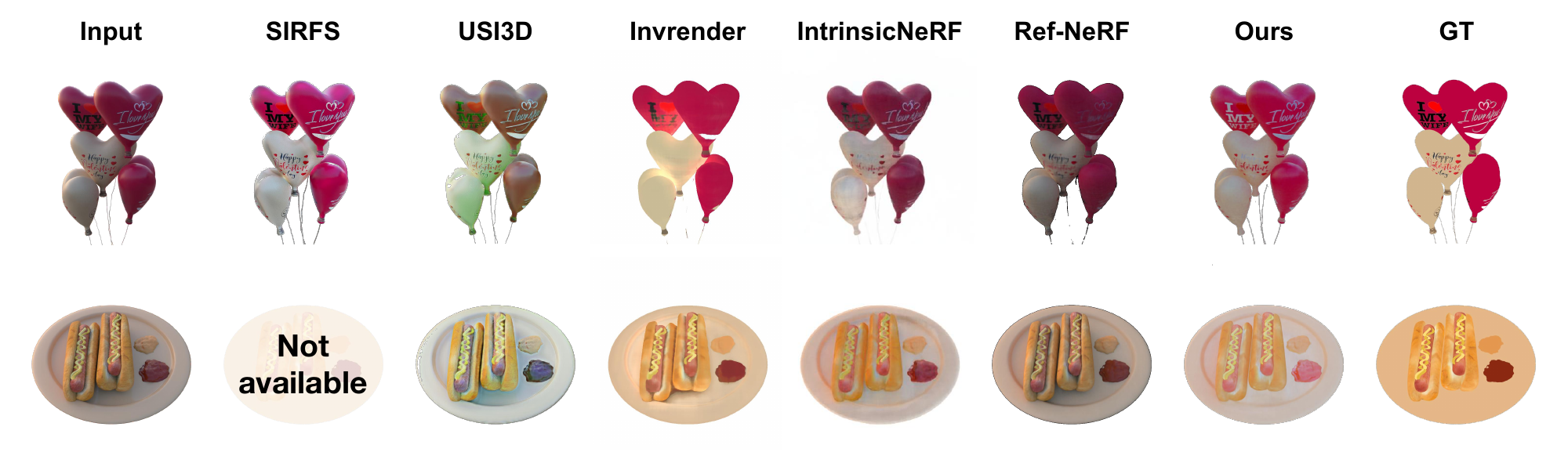}  
      \caption{\label{fig:exp1}%
      Reflectance decomposition results. SIRFS \cite{2015Shape} and USI3D \cite{Liu2020Unsupervised} are tested directly using pre-trained networks.%, and hotdog and jugs are not applicable in SIRFS. The 
      Invrender\cite{zhang2022invrender}, IntrinsicNeRF~\cite{ye2023intrinsicnerf}, Ref-NeRF~\cite{verbin2022ref}, and our method use the same training data. }
      % ``Ours'' denotes our method.  
      % ``Ours*'' denotes our method by reducing the weights shading smoothness and increasing the weights of reflectance sparsity. }%\color{red}{delete the bottom two data. }}%Column Our is our result, and column Our* is another decomposition result after adjusting the weight of the regularization constraints.}
      \vspace{-5px}
\end{figure*}
 \begin{figure}
 \vspace{-5px}
      \centering
\includegraphics[width=\linewidth]{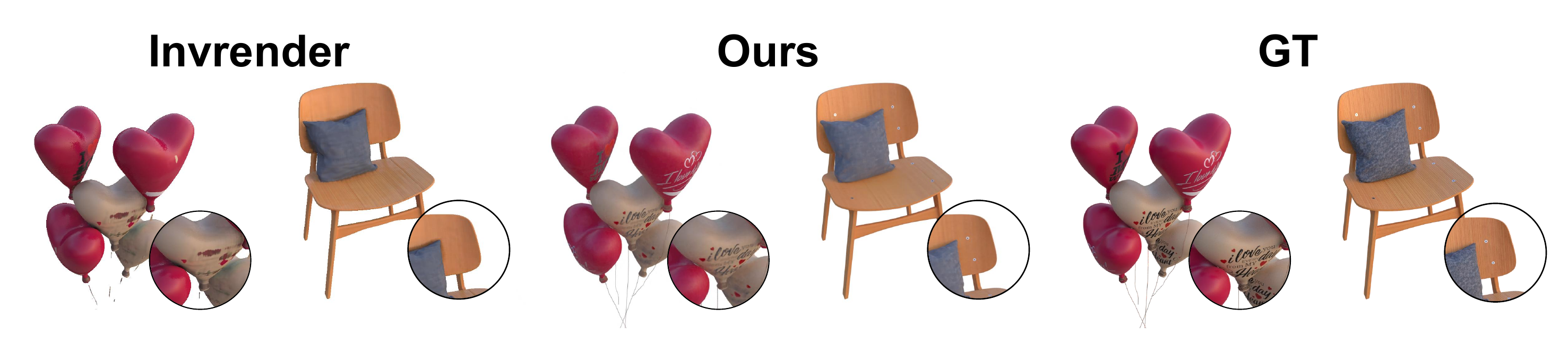}  
      \caption{\label{fig:exp-supp2}%
       New perspective synthesis results for objects. Our method achieves superior results in novel view synthesis compared to Invrender.}
        \vspace{-5px}
\end{figure}
\subsection{Implementation Details}
We adopt the codes of MipNeRF360 \cite{2021Mip} as the NeRF backbone and implement the method within the JAX framework \cite{jax2018github}. 
Our model was trained on an RTX 3080 Ti GPU using the Adam optimizer \cite{kingma2014adam} for a total of 100 thousand iterations. 
Throughout the training process, one image was randomly sampled from the input images per iteration, with a batch size set to 1024.
To facilitate the acquisition of neighboring pixels $\mathbf{r}_{adj}$ and non-neighbor pixels $\mathbf{r'}$, we adopt distance-aware sampling from \cite{ye2023intrinsicnerf}. 
In all experiments, we set parameters as 
$\lambda_0=1$, $\lambda_1=0.01$, $\lambda_2=0.1$, $\lambda_3=1$, $\lambda_4=0.01$, $\lambda_5=0.005$, $\lambda_6=0.1$, $\lambda_7=0.001$. 
The evaluations leverage MipNeRF360 datasets \cite{2021Mip}, Blender datasets \cite{2020NeRF}, and Invrender datasets \cite{zhang2022invrender}. 
The evaluation metrics include Peak Signal-to-Noise Ratio (PSNR), Mean Squared Error (MSE), Structural Similarity Index Measure
(SSIM) \cite{2013Image} and Learned Perceptual Image Patch Similarity (LPIPS) \cite{zhang2018perceptual}.

%\vspace{-6px}

\subsection{Comparsions}% and Applications}
%We evaluate intrinsic decomposition, new view synthesis, composition, and relighting. We also conduct several ablation studies. 
%The results are divided into four parts. 
%The first part is the comparison of the intrinsic decomposition,
%the second part is the ablation studies,
%the third part is the composition of the object and scene, 
%the fourth part is the relighting result. 
%In the next section, we also show applications such as re-colorization. % and the last part is for other applications.

\subsubsection{Intrinsic Decomposition}

%In this part of the experiment, we evaluate the ability of Intrinsic Neural Radiance Field decomposition. 
Although the proposed method decomposes intrinsic NeRFs, we also compare them with intrinsic image decomposition baselines, which take a single image as input. 
%Due to our decomposition method being based on the intrinsic image decomposition algorithm, 
We compare with USI3D~\cite{Liu2020Unsupervised} and SIRFS~\cite{2015Shape}, which solve the intrinsic image decomposition from a single picture. We also compare with Invrender~\cite{zhang2022invrender}, Ref-NeRF~\cite{verbin2022ref} and IntrinsicNeRF~\cite{ye2023intrinsicnerf}, which combine NeRF with inverse rendering decomposition, taking multi-view images as inputs.  
%Due to the lack of released codes for IntrinsicNerf~\cite{ye2023intrinsicnerf}, we only compare with their results on selected data from their paper. 
Specifically, we use the dataset from Invrender~\cite{zhang2022invrender} and evaluate the decomposed reflectance. 
Fig.~\ref{fig:exp1} shows the reflectance images decomposed by our method and five other methods.
Table~\ref{table:exp1} presents the quantitative evaluation results.
The decomposition results based on a single image are not satisfactory due to the lack of information from multiple perspective images. They cannot decompose correct reflectance colors due to ambiguities. 
Methods from multi-view input can fully utilize cross-view information and successfully decompose better reflectance colors.
SIRFS \cite{2015Shape} fails to decompose in hotdog data. 
%{\color{red}{and IntrinsicNerf did not show this data in their paper. }}
Invrender~\cite{zhang2022invrender} achieves the best performance on error metrics, but their reflectance loses many details, such as the texts on the balloons in Fig.~\ref{fig:exp-supp2}. 
They also have difficulties in predicting objects with complicated geometry.
IntrinsicNeRF~\cite{ye2023intrinsicnerf} get comparable results with us by extra semantic information.
% as stated in . %since they reconstruct geometry based on the IDR method~\cite{yariv2020multiview}. 
Ref-NeRF~\cite{verbin2022ref} fails to effectively separate the reflectance and geometry information, particularly for objects with Lambertian surfaces.

\begin{table}[!ht]
      \centering
      \scalebox{1}{
      \begin{tabular}{lllll}
      \hline 
            ~ & MSE↓ & PSNR↑  & SSIM↑  & LPIPS↓ \\ \hline
           USI3D  & 0.0097  & 21.8434 & 0.9263  & 0.1395  \\ 
           SIRFS  & 0.0089  & 20.6065  & 0.9294  & 0.1097\\
           Invrender & \textbf{0.0019}  & \textbf{28.1937}  & \textbf{0.9336}  & \textbf{0.0764}  \\ 
           IntrinsicNeRF   &0.0038 & 25.1896  &0.9221    &0.0871   \\ 
           Ref-NeRF  & 0.0042  & 24.5182  & 0.9186 & 0.0985  \\ 
          % ~ & Ours & \underline{0.0040} & 25.2196 & 0.9270 & 0.1037 \\ 
           Ours & \underline{0.0037} &\underline{25.6568} & \underline{0.9314} &\underline{0.0784} \\ \hline
      \end{tabular}
      }
      % \vspace{2mm}
      \caption{Quantitative evaluations of decomposed reflectance.
      %All methods are tested on {\color{red}{four sets of data}}, except for SIRFS~\cite{2015Shape}, which failed on two data. So here we evaluate it on two sets of experiments. %We use the scale-invariant MSE to align the reflectance before conducting numerical evaluation due to the scale ambiguity in intrinsic image decomposition. 
      The best values are in bold, and the second-best values are underlined. }%This method was also used in the SIRFS.}
      \label{table:exp1}
% }
\vspace{-8px}
\end{table}

\begin{table}[!ht]
      \centering
      \scalebox{1}{
      \begin{tabular}{lllll}
      \hline 
           ~ & MSE↓ & PSNR↑  & SSIM↑  & LPIPS↓ \\ \hline
           Invrender & 0.0011  & 29.7887  & 0.9461  & 0.0809  \\  
           Ours & \textbf{0.0004} & \textbf{34.9118} & \textbf{0.9675} & \textbf{0.0346} \\ \hline
      \end{tabular}  
      }
      % \vspace{2mm}
      \caption{Quantitative evaluations for novel view synthesis. %The experiment settings are the same as in Intrinsic decomposition. 
    }
      \label{table:exp2}
      % \vspace{-15px}
       \vspace{-10px}
\end{table}

\subsubsection{Novel View Synthesis}
For novel view synthesis, we compare our results with Invrender~\cite{ye2023intrinsicnerf}.
% and MipNeRF360~\cite{2021Mip}.
The results are shown in Fig.~\ref{fig:exp-supp2}. 
As we can see in Fig.~\ref{fig:exp-supp2}, the images rendered by Invrender lose the original details of the object, such as the text on the balloon and the round metal screws on the back of the chair. They cannot handle objects with small or complex textures very well. Our method surpasses Invrender across all metrics, with the results presented in Table \ref{table:exp2}.
% For scene data, we conduct comparative experiments with MipNeRF360.
% %Because our method focuses more on the foreground of the scene, we can
% Even though our method reconstructs each image by reflectance and shading during rendering, we still generate more details of the scene. For example, the lines on the table in the garden scene and the white branch on the left front in the stump in Fig.~\ref{fig:exp-supp3}. %For object data, 

% \begin{figure}
%       \centering
% \includegraphics[width=\linewidth]{fig2/exp1-supp3.pdf}   
%       \caption{\label{fig:exp-supp3}
%        New perspective synthesis results for scenes. We compare our model with MipNeRF360 with MipNeRF360 datasets. }%\color{red}{make it into one row}}
% \end{figure}

\begin{figure*}
      \centering
        \includegraphics[width=\linewidth]{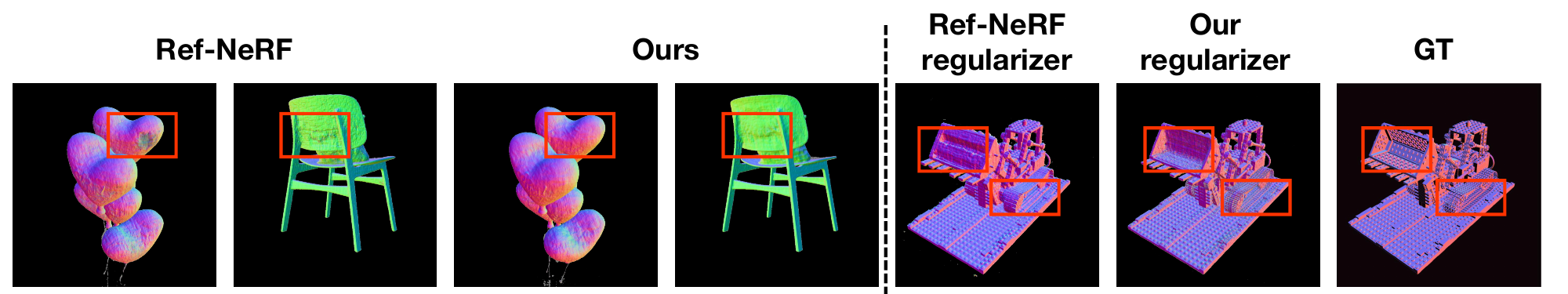}   
      \caption{\label{fig:ref-normal}%
     Normal comparisons with Ref-NeRF. The left part shows the normal reconstruction results of Ref-NeRF and our method. The right part compares the normal reconstruction with the regularizer of Ref-NeRF and ours.}
     % \vspace{-2mm}
\end{figure*}
\begin{figure*}
      \centering
      \includegraphics[width=\linewidth]{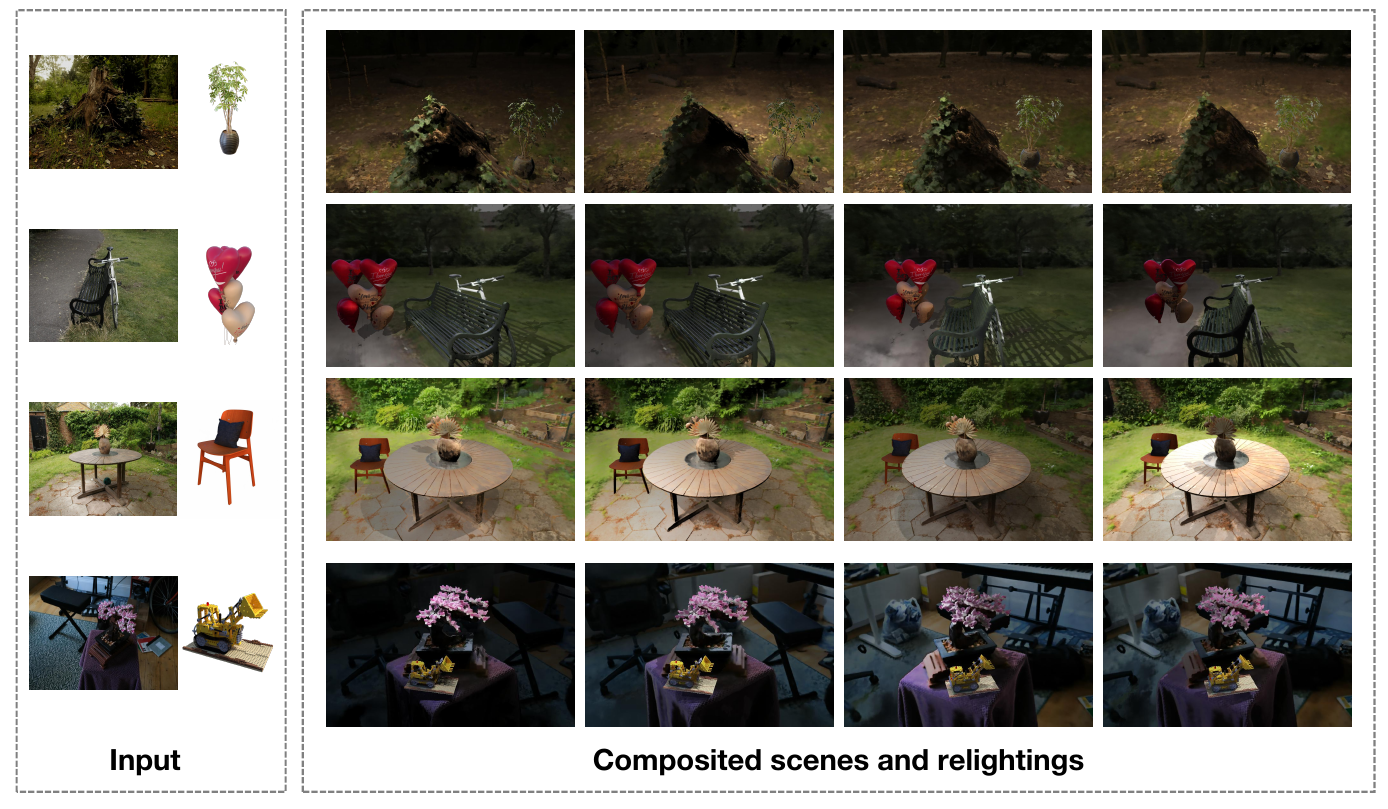}   
      \caption{\label{fig:exp2}%
      Composition and relighting results of four scene-object pairs. %of scene and object composition. 
      For each composited scene, we relight them under 4 different lighting.}%All results are with shadows. For each set of data, we show two views (``view \#1'', ``view \#2'') \textcolor{red}{and three lights (``light \#1'', ``light \#2'', ``light \#3'').}}
      \vspace{-8px}
\end{figure*}

\subsubsection{Normal Reconstruction}

The normal estimation of our method is based on the NeRF backbone. We compare our method with Ref-NeRF \cite{verbin2022ref}, another NeRF-based method.  
The dataset of Invrender was reconstructed on Ref-NeRF and our method.
The left part of Fig.~\ref{fig:ref-normal} shows the reconstruction results of Ref-NeRF.
It is observed that the surface reconstructed by our method is smoother such as the back of the chair and the surface of the balloon. 
This is because we adopt the constraint of normal smoothing and restrict the consistency between normal and shading, resulting in a reasonable physical connection between geometry and lighting.
%Therefore, our reconstruction results are better and more reasonable.
\begin{figure*}
      \centering
      \includegraphics[width=\linewidth]{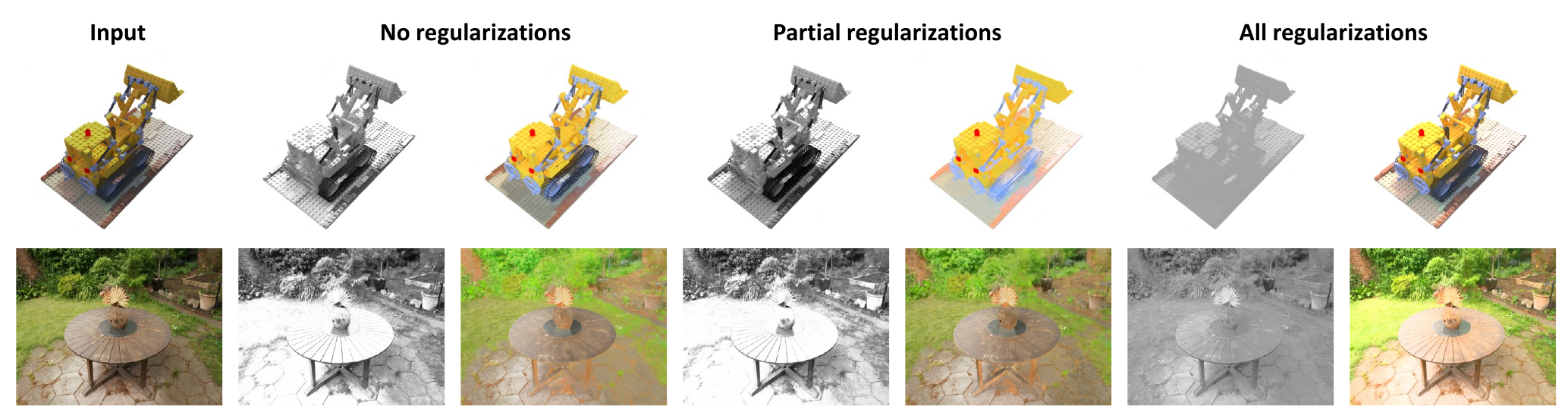}   
      \caption{\label{fig:ablatio0}%
      Ablation studies on regularization constraints. ``No regularizations'' denotes trained with only RGB loss. ``Partial regularizations'' denotes constraints of IntrinsicNeRF \cite{ye2023intrinsicnerf} ($\mathcal{L} _ {chro}$, $\mathcal{L} _ {rs}$, $\mathcal{L} _ {rns}$). }
      %``All'' denotes all our constraints. }
\end{figure*}

\begin{figure*}
      \centering
\includegraphics[width=\linewidth]{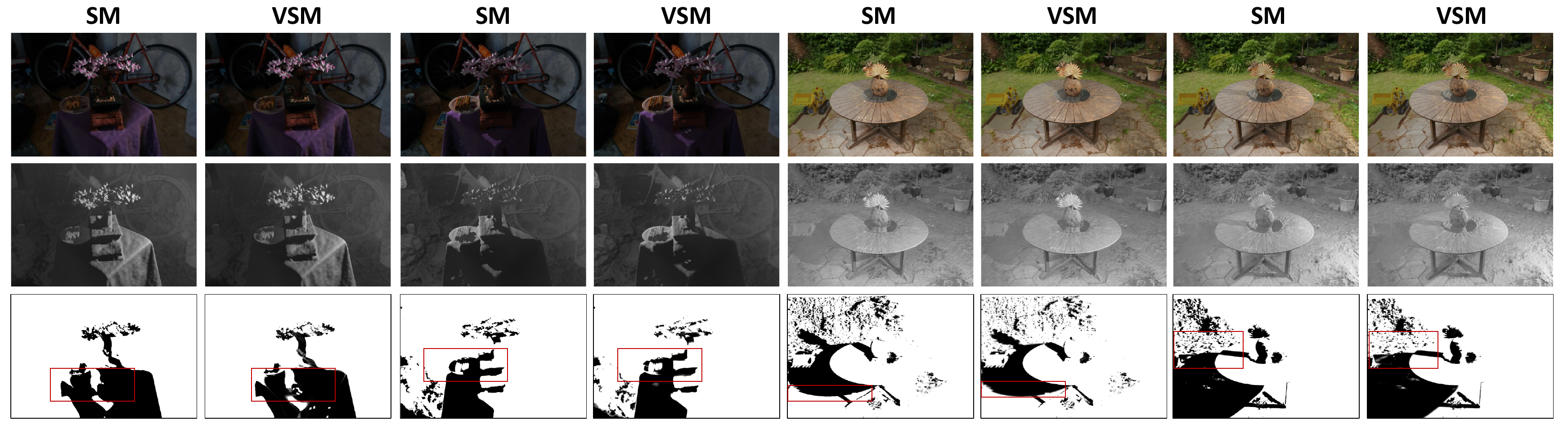}   
      \caption{\label{fig:shadowmapping}%
      Comparisons between variance shadow mapping (VSM) and naive shadow mapping (SM). The odd and even columns are paired for comparison. From top to bottom, three rows are rendered images, SG shadings, and cast shadows. VSM can model soft shadows while SM models only hard shadows. %The first and second lines are under the same lighting condition, while the third and fourth lines are also under the same lighting condition. The first three lines are a hotdog inserted into a bonsai scene, while the last three lines showcase Lego inserted into a garden scene.
      }
\end{figure*}

% \vspace{-6px}

\subsection{Scene Composition and Relighting}%of Object and Scene}

% \begin{figure*}
%       \centering
%       \includegraphics[width=\linewidth]{exp3.pdf}   
%       \caption{\label{fig:exp3}%
%       In the relighting experiment of a single object, shadows are added to all results.
%       Ground truth (GT) images are obtained by Blender Cycles. Each object is tested for two views (view \#1, view \#2) and two lights (light \#1, light \#2).}
% \end{figure*}

% \begin{figure*}
%       \centering
%       \includegraphics[width=\linewidth]{exp4.pdf}   
%       \caption{\label{fig:exp4}%
%       Application of re-colorization. By modifying the reflectance NeRF of the object, the effect of modifying the color of the object is achieved. Different from methods based on image editing, we can render the recolored scene from any perspective.  }
% \end{figure*}

For composition, we use the MipNeRF360 dataset \cite{2021Mip} as scene data, the Blender dataset \cite{2020NeRF}, and the Invrender dataset \cite{zhang2022invrender} as object data.
We use four different scene and object pairs in experiments.
%The scene image used for training is four times the size of the original downsampling.
% We show experiments on inserting objects into scenes, as shown 
In Fig.~\ref{fig:exp2}, The input scene and object pairs are shown on the left, and relightings by three different lighting conditions from different views are shown on the right.

% The first two rows are relit by adding new lights on the scene lighting, and the last row shows replacing the scene lighting with a point light. 
% removes the original scene lighting and directly adds a new light source.}}
%we show images of adding {\color{red}{two ?}} different lights from two perspectives each. 
%We randomly insert objects into the scene and add light sources to achieve a new perspective image with different lighting.
%{\color{red}{The second and third columns represent the effects of different lighting from the same perspective, while the fourth and fifth columns also represent the effects of different lighting from another perspective. The lighting in the second and fourth columns is the same, and the lighting in the third and fifth columns is the same.}}
Through efficient shadow addition, we have achieved the effect of scene and object interaction of casting shadows on each other. %, such as the shadow of the tree stump in the first row of Figure \ref{fig:exp2} falls on the ground, and the shadow of the flower pot also falls on the ground of the scene. The shadow of the second row of chairs fell on the ground and balloons. The shadow of the third-row table falls on the chair. The shadow of Lego in the fourth line falls on the table.
For example, in the last column, the shadow of the plant is cast onto the Lego. 
%All objects are inserted into the scene by our efficient shadow addition modules, with the light and shadow interactions between the inserted objects and the scene, making the overall image more realistic. 
%In the supplementary video, we show more video demos of relighting, as well as novel view synthesis. We show videos of rotation lighting and rotation views. 
The proposed method supports relighting from arbitrary views. 
%It shows that object insertion by our method enables relighting and view synthesis at the same time. 
Previous methods of single-image object insertions can only relight the single-view image. 
\vspace{-5px}
\subsection{Ablations}\label{sec:ablation}
To verify the effectiveness of our method, we conducted several ablation studies. 
The scene data is sourced from MipNeRF360 \cite{2021Mip}, and the object data is from NeRF \cite{2020NeRF}. 
Ablations of ``decomposition with and without regularization constraints'', ``variance shadow mapping and naive shadow mapping'', ``shading replacement'' and ``hybrid parametric lighting model'' are introduced below.

\begin{figure*}
      \centering
      \includegraphics[width=\linewidth]{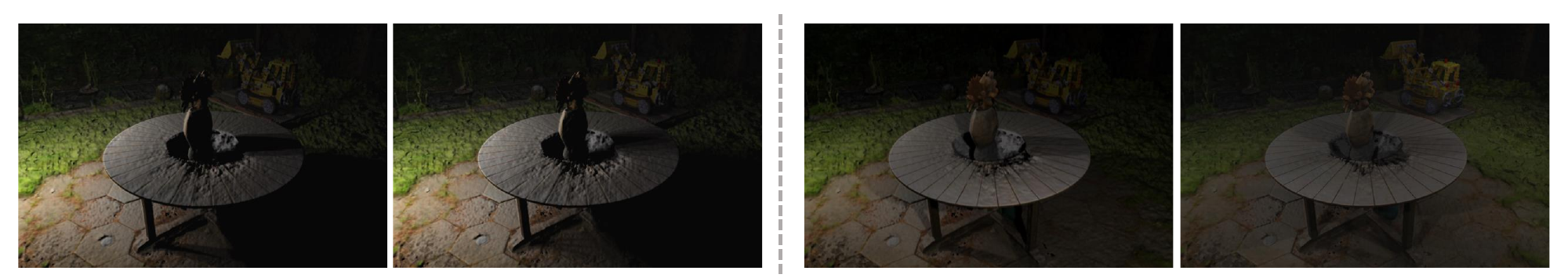}   
      \caption{\label{fig:soft_shadow}%
      Relighting with nearby (left) and faraway (right) SG lighting, producing soft shadows (left) or sparse shadows (right). }
\end{figure*}
\subsubsection{Regularization Loss} 
Since the unsupervised intrinsic image decomposition problem is ill-posed, we add some regularization constraints.
We conduct ablation studies on these regularization losses. We test three settings, without adding any regularization losses, using partial regularization losses ($\mathcal{L}_{chro}$, $\mathcal{L}_{rs}$, $\mathcal{L}_{rns}$), and using all regularization losses.
Fig.~\ref{fig:ablatio0} shows the comparison results of the three settings. 
Without any regularization constraints, reflectance and shading decomposition are relatively arbitrary. Both are not very accurate. 
After using some regularization from IntrinsicNeRF \cite{ye2023intrinsicnerf}, the reflectance becomes more flattened, but the shading is unsatisfactory. The shading contains too much color information belonging to reflectance. With all regularization constraints, it decomposes the shading image much better, not including surface colors. The decoupling effect of shading results in a better and more accurate representation of the lighting conditions. For example, in the case of the Lego wheel, the shading generated by our method does not include the color of the wheel. Instead, the black color of the wheel, derived from partial regularization, is still black in the shading. 

Considering that the Ref-NeRF~\cite{verbin2022ref} also adopts normal regularization, %to compare the effectiveness of regularizers more fairly, 
we replace our normal regularizer with the Ref-NeRF regularizer in our method.
The result of the comparison between the two regularizers is shown in the right part of Fig.~\ref{fig:ref-normal}.
%PSNR is $20.47$ for ours and $19.98$ for the Ref-NeRF regularizer. 
Qualitatively, the geometry by Ref-NeRF regularizer is over-smoothed in small structures, such as the wheels and shovel of Lego (in the red box in Fig.~\ref{fig:ref-normal}). 

\begin{figure}
      \centering
      \includegraphics[width=\linewidth]{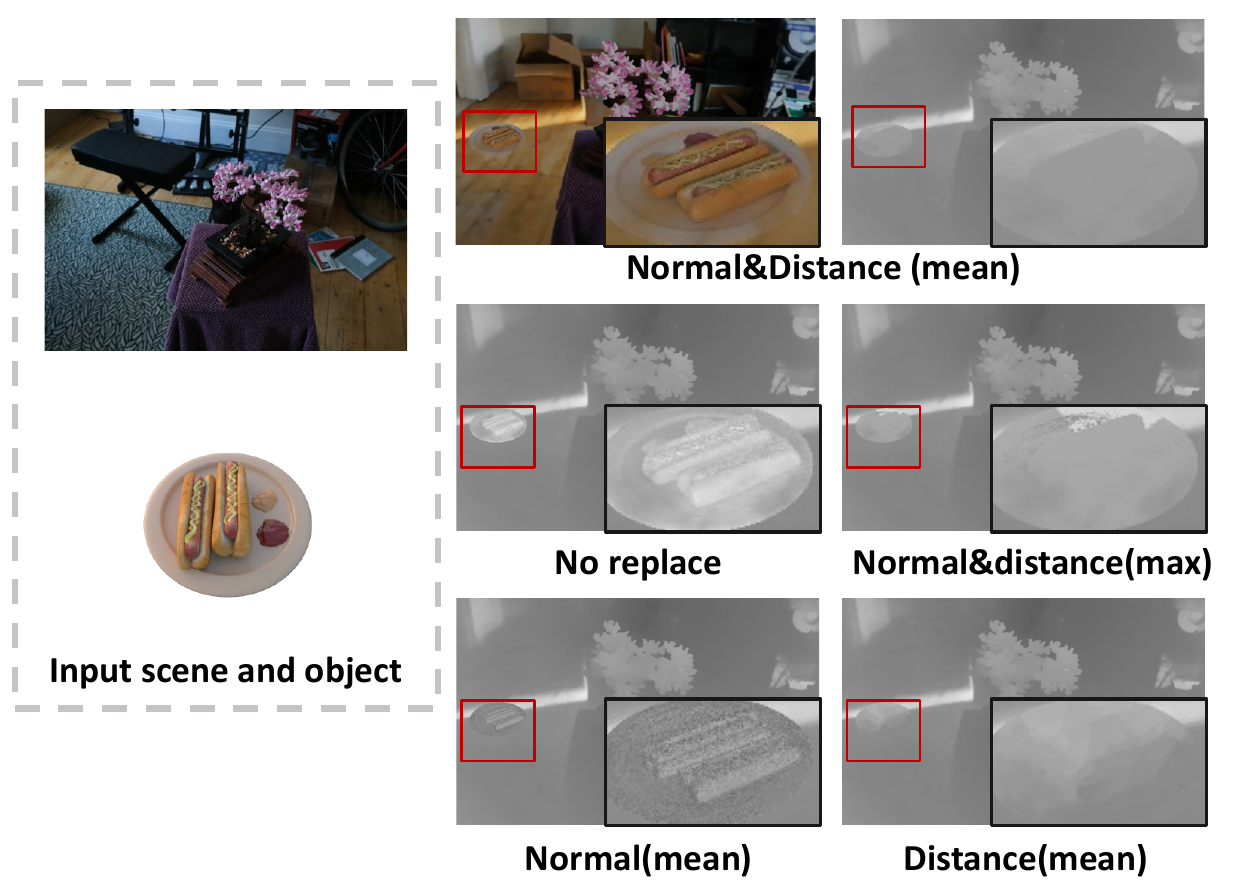}   
      \caption{\label{fig:ablatio1}%
        Ablation experiments on shading replacement. At the left side, the first row shows the shadow replacing strategy adopted in the method, and the last two rows show other alternatives. }%our result and the second row shows the other shading results.}
\end{figure}

% \begin{figure}
%       \centering
% \includegraphics[width=\linewidth]{blender2.pdf}   
%       \caption{\label{fig:evaluteRelighting}%
%       Relighting results. The results of ours and GT. }
% \end{figure}  

\subsubsection{Variance Shadow Mapping vs Naive Shadow Mapping}
%There are many algorithms for shadow mapping. Therefore, 
We compare naive shadow mapping with variance shadow mapping to demonstrate the soft shadows. In naive shadow mapping,
due to the lack of handling abrupt changes in depth in regular shadow mapping, the resulting shadows are discrete values of 0 and 1 and the shadow edges are abrupt and sharp. It can only produce hard shadows. Variance shadow mapping smoothens the depth and calculates the probability of shadows using Chebyshev's inequality, leading to smoother shadow edges. As illustrated in Fig.~\ref{fig:shadowmapping}, at the edges of the shadows, like the shadow edge of a table on the ground in a garden scene or the shadow edge of a plate in a bonsai scene, there is a gradient effect, creating a soft shadow effect.
Additionally,  variance shadow mapping requires storing two depth maps and filtering them, resulting in only slightly increased computational overhead compared to shadow mapping. %However, it achieves the effect of soft shadows. 
Therefore, we select variance shadow mapping as the method for shadow generation. When the SGs are near from each other, it generate soft shadows as in Fig.~\ref{fig:soft_shadow} (left). When SGs are far from each other as in Fig.~\ref{fig:soft_shadow} (right), the shadows are also sparsely distributed, similar in the real cases.

\subsubsection{Shading Replacement}
Due to substantial lighting variations, relighting objects after insertion into a scene becomes crucial. This relighting is achieved through shading replacement, avoiding the need to infer scene illumination.
%To demonstrate the effectiveness, we compare the results with and without the shading replacement step. % replacing shading and with replacing shading. 
By specifying a confidence level $p$, the top $p$ pairs are selected based on a scoring system considering factors such as normal similarity, distance score, and their product. 
% When only based on normal similarity, shading is obtained globally. 
% When only based on distance, shading does not conform to the physical meaning. 
Solely relying on normal similarity results in global shading, while focusing solely on distance fails to align with physical interpretations.
% When using the product of the two, it is possible to achieve query shading within a local range.
However, by combining both factors, it becomes feasible to attain query shading within a localized range. 
On the other hand, we compare two strategies, taking the average and maximum shading of $p$ points.
As in Fig.~\ref{fig:ablatio1}, 
directly inserting a hot dog into the scene leads to an unrealistic composition, as the original lighting environment of the hot dog is different from the scene lighting. %, resulting in the Lego part being too bright in the combined result. 
When we replace the shading of the hot dog, the composition of the hot dog and the scene are more consistent. 
Replacing shading with normal similarity results in too much noise. 
Replacing shading with maximum values cannot generate plausible shading.
Using mean shading gives a reasonable result without introducing noises. %, resulting in a high amount of shading noise, resulting in many noise points in the combined Lego, as shown in the last two columns of Figure \ref{fig:ablatio1}.

\begin{figure}
      \centering
\includegraphics[width=\linewidth]{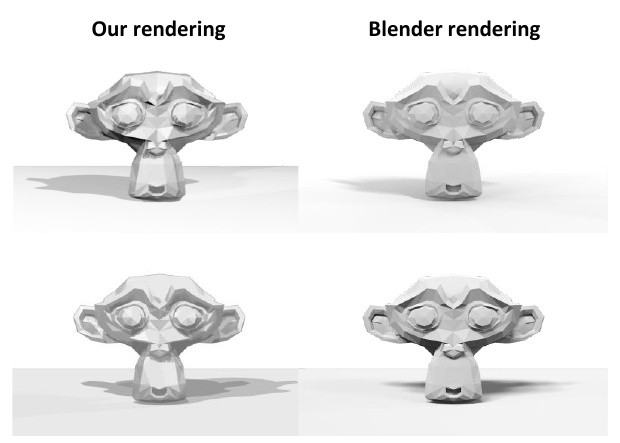}   
      \caption{\label{fig:evaluteRelighting}%
      Relighting by the proposed renderer and Blender. }% results. The results of ours and GT. }
\end{figure}  

\begin{figure*}
      \centering
\includegraphics[width=\linewidth]{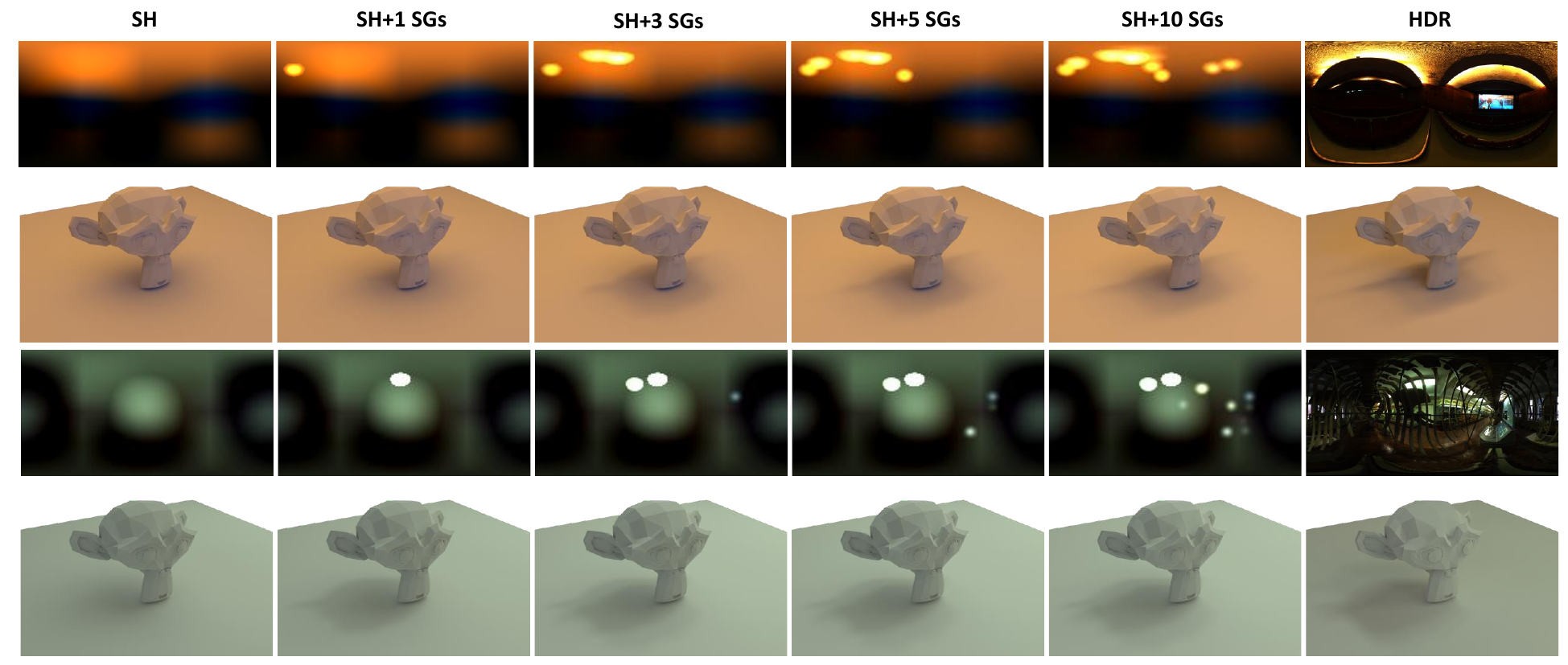}   
      \caption{\label{fig:sh+sg}%
      Hybrid parametric lighting model. The render results of the SH lighting model, hybrid parametric lighting model, and GT. The first and third rows are spherical lighting maps. The second and fourth rows are corresponding renderings. }
\end{figure*}

\begin{figure}
      \centering
      \includegraphics[width=\linewidth]{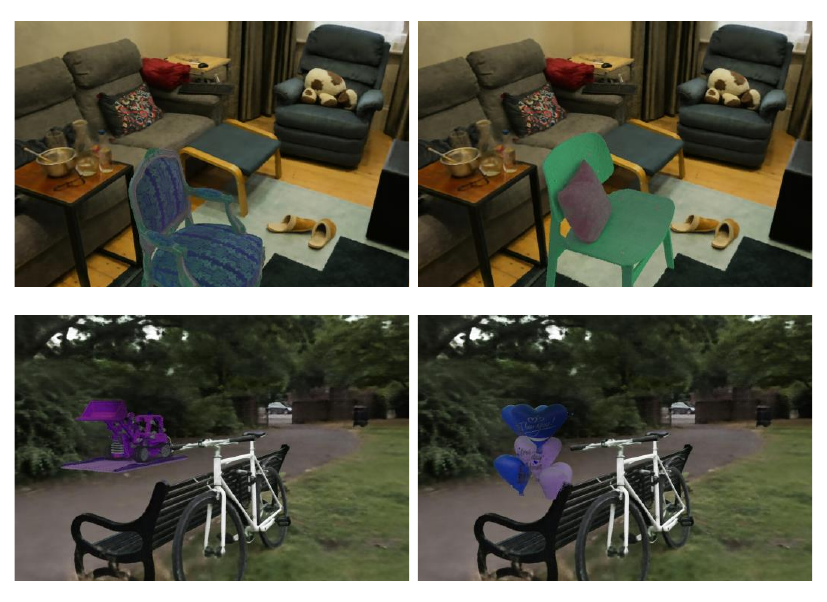}   
      \caption{\label{fig:exp4}%
      Demonstration of the re-colorization effects. The reflectance of the original objects are changed after intrinsic decomposition. } %Application. The first two rows are the results of changing color, the last two rows are the results of changing material. }
      \vspace{-5px}
\end{figure}
\subsubsection{Hybrid Parametric Lighting Model}
we conduct ablation experiments on the hybrid parametric lighting model and the standalone SH lighting model, as well as ablation experiments with different numbers of SG.
Specifically, we render the original HDR, SH representation and SH+SG representation in Blender, to compare the accuracy of the hybrid lighting representation. %perform rendering experiments in Blender using HDR images generated with SH and HDR images generated with our hybrid parametric lighting model.
As in Fig.~\ref{fig:sh+sg}, Spherical Harmonics (SH) is capable of fitting only low-frequency lighting, and the resulting renderings lack distinct shadows. Our hybrid model was able to produce noticeable shadows, showcasing its ability to capture both low- and high-frequency lighting in original HDRs.
Additionally, we found that increasing the number of SGs can bring the results closer to ground truth (HDR), while the computational costs also increase. We observe that when the number of SGs reached 3, the experimental outcomes are already sufficiently representative of the real scenario, even for indoor scenes. Outdoor scenes are even simpler with one dominant lighting source (sun or moon). Further increasing the number of SGs does not significantly alter the results. So we set the number of SGs as 3 in all relighting experiments. 

We also compare the rendering of our pipeline with Blender in Figure~\ref{fig:evaluteRelighting}. While our renderer is much more efficient, the rendered images and shadows are similar, while Blender still generates more soft shadow edges. % and the 

% \begin{figure*}
%       \centering
%       \includegraphics[width=\linewidth]{video.pdf}   
%       \caption{\label{fig:relighting1}%
%       More results of relighting. The view is fixed and the lighting is changed.
%       The light source(point light) moves from the left rear to the left front, gradually towards the right rear.}
% \end{figure*}

% Fig.~\ref{fig:relighting1} shows more results of relighting, the result and new shading after relighting can be seen. As the light source moves around in a full circle, the shading clearly showcases the effects of the changing light source position, which is also reflected in the final image.

\begin{figure}
      \centering
\includegraphics[width=\linewidth]{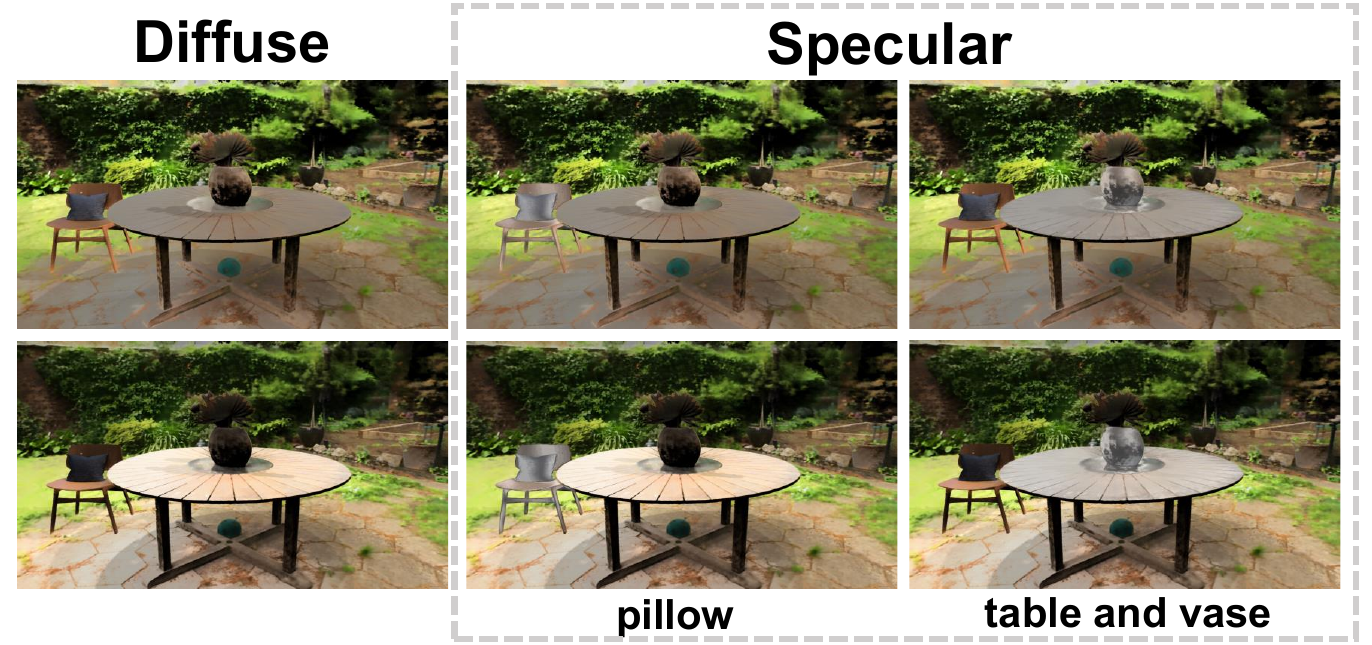}   
      \caption{\label{fig:spcular}%
      Demonstration of re-material effects. The first row shows all-diffuse results, while the subsequent rows demonstrate re-material results of pillows and vases respectively. }
\end{figure}

% \begin{figure}
%       \centering
%       \includegraphics[width=0.95\linewidth]{morelights4.pdf}   
%       \caption{\label{fig:morelights_4}%
%       Relighting with multiple(4) light sources. The density of light source distribution is different for each row, with higher row numbers corresponding to higher densities.}
% \end{figure}

\vspace{-5px}

\subsection{Application}

The proposed method has many applications, except for scene composition and relighting. 
Here we show the effect of changing the color of objects by modifying their reflectance in NeRFs. 
As shown in Fig.~\ref{fig:exp4}, we insert 2 different chairs into the room scene and modify them to different colors. In the first row, purple Lego and blue balloons are inserted into the bike scene. %Different from methods based on image editing, we can render the edited scene from any perspective.
In Fig.~\ref{fig:spcular}, we change the material of the pillow, table, and vase of the original scene by making them more specular. Compared to the all-diffuse scene, we can edit the material of each part of the composited scene.

%-------------------------------------------------------------------------
% \vspace{-6px}
\section{Conclusions}
We propose a method for compositing objects and scenes in NeRFs. The object insertion considers the lighting consistency in the composited scene, or relighting the whole scene. We use a hybrid representation of Spherical Harmonics and Spherical Gaussians, to support non-Lambertian rendering with shadows. One limitation is manually setting an insertion position instead of automatically attaching to surfaces. Ideally, automatically attached to nearby surfaces would be more practical in AR applications. We will further explore such physical interactions. 
%Specifically, taking two sets of posed images of the object and the scene as inputs, the proposed method achieves the realistic combination of two NeRFs.
%In intrinsic image decomposition based on Neural Radiance Fields, we introduce regularization constraints to guide decomposition to deal with this ill-posed problem.
%We achieve the insertion of objects into the scene and make the image appear more realistic by adjusting the lighting of the object parts.
%By the proposed efficient shadow rendering, we can add light sources to adjust the lighting condition of the composited scene. The proposed shadow rendering method avoids the extensive integration calculation along the shadow rays. 
%One limitation is the current method cannot deal with non-Lambertian surfaces. For non-Lambertian scenes, the strategy of shading replacement is infeasible, which will be further explored as future work. 
%Another limitation is the lack of low-frequency environment lighting models, current point light rendering may lead to extensive computational and time costs if the new light sources are combinations of area lights or ambient lights. Using parametric lighting models such as Spherical Harmonics or Spherical Gaussians would be much more efficient in rendering. 